\documentclass[10pt,twocolumn,letterpaper]{article}

\usepackage[pagenumbers]{cvpr} 

\definecolor{cvprblue}{rgb}{0.21,0.49,0.74}
\usepackage[pagebackref,breaklinks,colorlinks,allcolors=cvprblue]{hyperref}

\usepackage{kotex}
\usepackage[dvipsnames]{xcolor}
\definecolor{mygreen}{HTML}{009700} 
\usepackage{url}
\usepackage{tcolorbox}
\usepackage{algorithm}
\usepackage{algpseudocode}
\usepackage{booktabs, tabularx}
\usepackage{multirow}      
\usepackage{makecell}  
\usepackage{amsmath,amssymb}
\usepackage{amsthm}        
\usepackage{hyperref}      
\usepackage{courier}       

\theoremstyle{remark}

\usepackage{xspace}
\usepackage{graphicx}
\usepackage{xcolor}
\usepackage{colortbl}
\usepackage{wrapfig}
\usepackage{graphicx}
\usepackage{fontawesome5}
\usepackage{subcaption}
\usepackage{xcolor}
\usepackage{etoc}

\definecolor{codecrimson}{HTML}{862633}

\hypersetup{
    colorlinks,
    citecolor=cvprblue
}

\def\confName{CVPR}
\def\confYear{2027}

\title{CommitQuant: Commitment-Aware Quantization \\ for Multimodal Diffusion Language Models}

\author{
Jaehyuk Yeon\textsuperscript{1}\quad
Donghyun Lee\textsuperscript{2}\quad
Zhuowei Li\textsuperscript{3}\quad
Linghan Xu\textsuperscript{4,\textdagger}\quad
Xi Xiao\textsuperscript{5}\\
Kyujin Shim\textsuperscript{6}\quad
Taeho Hwang\textsuperscript{6}\quad
Sungeun Hong\textsuperscript{7}\quad
Youngeun Kim\textsuperscript{1,*}\\[0.5em]
\textsuperscript{1}Korea University\quad
\textsuperscript{2}University of Southern California\quad
\textsuperscript{3}Rutgers University\quad
\textsuperscript{4}Amazon\\
\textsuperscript{5}University of Alabama at Birmingham\quad
\textsuperscript{6}SK hynix\quad
\textsuperscript{7}Sungkyunkwan University\\[0.6em]
{\normalsize
\href{https://yeontachi.github.io/commitquant-page/}
{\textcolor{codecrimson}{\faGlobe\ Project Page}}
\quad
\href{https://github.com/yeontachi/CommitQuant}
{\textcolor{black}{\faGithub\ Code}}
}
}

\usepackage[bottom]{footmisc}
\begin{document}
\etocdepthtag.toc{main}
\raggedbottom
\maketitle

\iftoggle{cvprfinal}{%
  \begingroup
  \setlength{\footnotemargin}{0pt}
  \renewcommand{\thefootnote}{\fnsymbol{footnote}}
  \footnotetext[0]{First author's email: \texttt{yeonjjhh@gmail.com}}
  \footnotetext[1]{Corresponding author (email: \texttt{youngeunkim@korea.ac.kr})}
  \footnotetext[2]{Work done outside of Amazon.}
  \endgroup
}{}

\vspace{-4mm}

\begin{abstract}

Multimodal diffusion language models (MDLMs) incur substantial inference costs from long visual contexts and iterative denoising. Low-bit quantization can reduce inference costs, but its errors can alter position--token decisions (i.e., commitments), thereby changing subsequent denoising states.
Loss gradients used to estimate modality sensitivity in autoregressive multimodal quantization are insufficient to guarantee the stability of these commitments.
Our analysis shows that, within the same MDLM, visual-token features exhibit smaller loss-gradient magnitudes but greater quantization-induced commitment discrepancy than textual-token features.
This mismatch motivates \textbf{CommitQuant}, a two-stage post-training quantization framework that targets full-precision (FP) commitment preservation. Our proposed method first calibrates decoder blocks using reconstruction weighted by modality-specific commitment discrepancies. To further align the predictions that drive commitments, we then jointly refine quantization parameters across blocks by minimizing KL divergence between FP and quantized token distributions at masked response positions under identical inputs.
Across three MDLMs and seven benchmarks, CommitQuant outperforms evaluated W4A4 baselines on average, improving MMMU by up to 1.33 percentage points over FlatQuant and achieving a $1.73\times$ generation speedup over FP for LLaDA-V-8B on an NVIDIA A100 GPU.
\vspace{-4mm}
\end{abstract}
   
\section{Introduction}
\label{sec:intro}
\vspace{-1mm}

\begin{figure}[t]
    \centering
    \hspace{-5.4mm}
    \includegraphics[width=1.055\columnwidth]{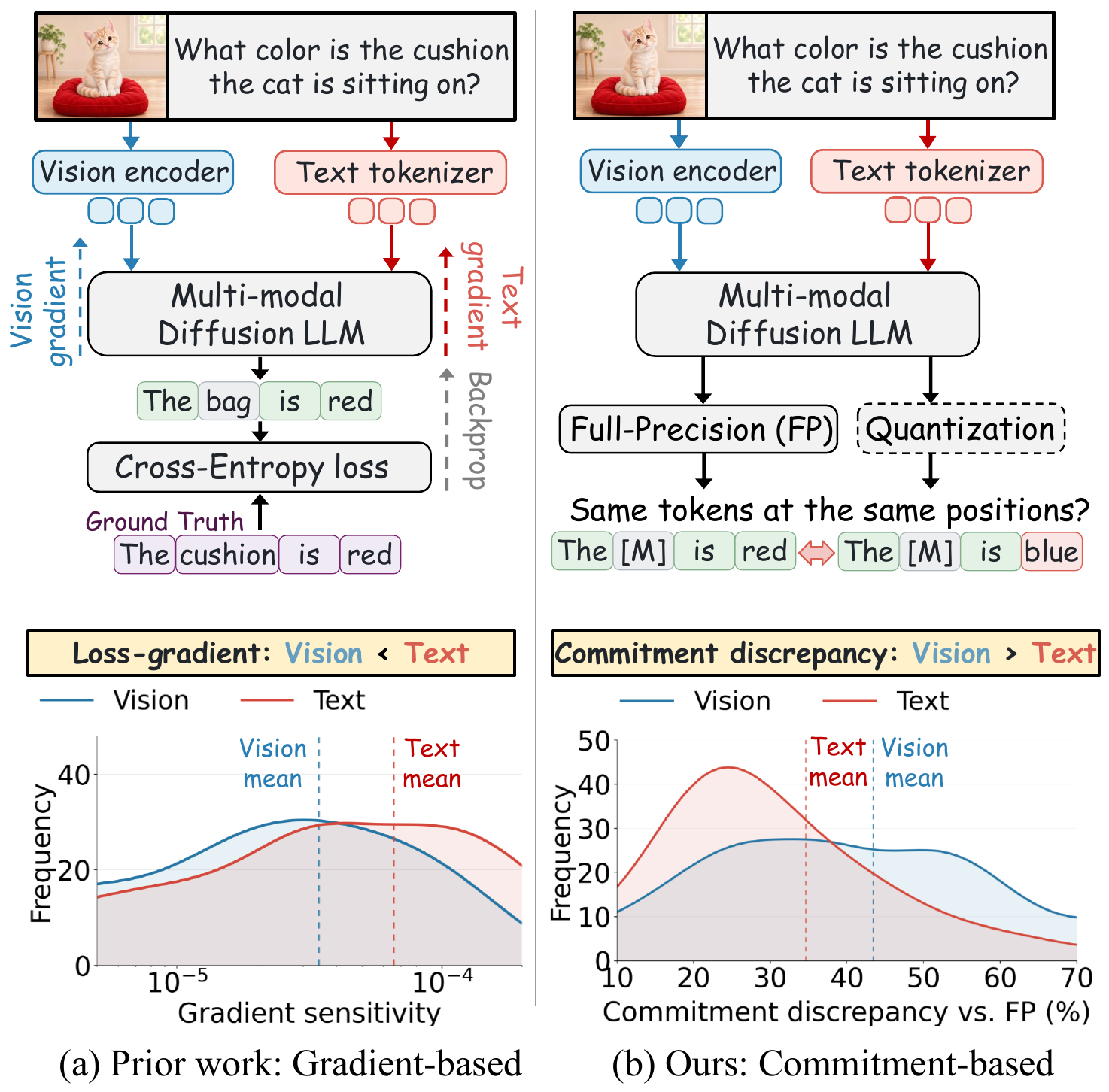}
    \vspace{-6mm}
 \caption{
\textbf{Modality sensitivity under loss-gradient and commitment-based measures.}
Both panels use LaViDa-LLaDA-8B and the same 256 TextVQA image--question pairs.
(a) Ground-truth cross-entropy at masked response positions yields larger mean gradient magnitudes for text-token features than for visual-token features.
(b) Visual-token errors induce greater mean commitment discrepancy than text-token errors. Discrepancy measures disagreement with FP in selected positions and assigned tokens, with visual and text errors injected separately on the same FP decoding states.
}
    \vspace{-2mm}
    \label{fig:motivation}
\end{figure}

Multimodal diffusion language models (MDLMs) generate responses from visual and textual inputs through iterative masked denoising~\cite{lavida, lladav, mmada}.
Starting from a masked output sequence, they predict tokens at masked positions and replace selected \texttt{[MASK]} tokens with these predictions at each step, progressively constructing the response.
Although this process supports parallel token prediction with bidirectional context, large parameter counts and repeated decoder evaluations impose substantial memory and computational costs~\cite{lavida, sparselavida, dllmquant}.
Low-bit quantization offers a practical route to reducing model storage and enabling lower-precision computation~\cite{mbq, mquant}. Despite progress in quantizing autoregressive vision-language models~\cite{qvlm, mbq, mquant, impquant} and text-only diffusion language models~\cite{qdlm, dllmquant, t2quant, soloq}, calibration strategies that account for modality-dependent quantization sensitivity during denoising remain underexplored in MDLM quantization.

In MDLMs, visual and textual representations jointly guide which masked positions are selected and which tokens are assigned to masked positions.
We refer to each position--token decision as a \emph{commitment}~\cite{shlomi2026spot}.
Quantization errors in visual or textual representations can change these commitments, thereby modifying the response context available to subsequent steps and potentially diverting generation from the full-precision (FP) trajectory~\cite{faircalib}.
Our objective is therefore to preserve FP commitments as closely as possible throughout denoising. For multimodal calibration, this raises a central question: \textit{which modality's representations require greater protection against quantization errors to preserve FP commitments?}

\begin{figure}[t]
    \centering
    \includegraphics[width=0.99\columnwidth]{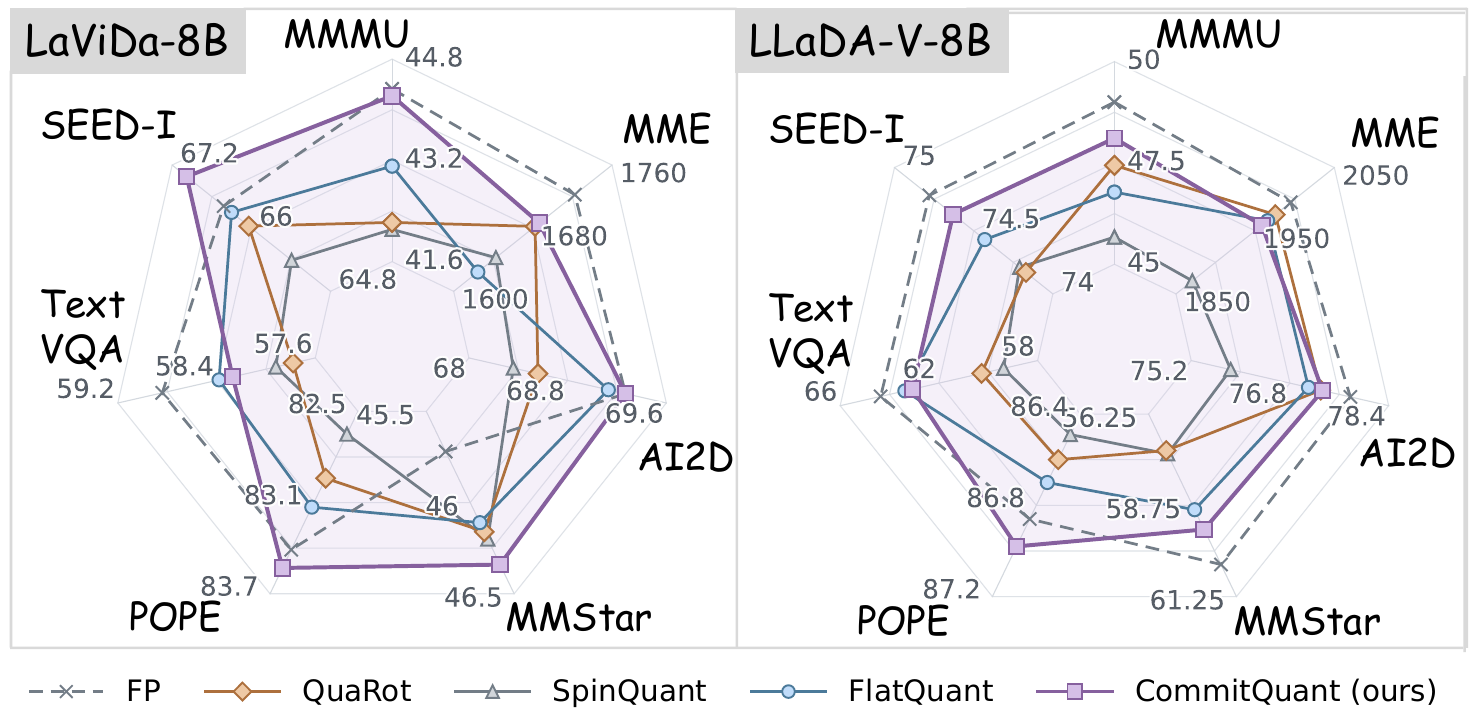}
    \vspace{-3mm}
 \caption{
\textbf{Results on LaViDa-LLaDA-8B (left) and LLaDA-V-8B (right)}. Quantized methods use W4A4, with full precision (FP) shown as a reference. Each axis displays raw benchmark scores over an independent linear range; higher is better.
}
    \vspace{-2mm}
    \label{fig:perf_overview}
\end{figure}

To answer this question, we assess how strongly quantization errors in each modality disrupt FP commitments, using their relative impact to guide reconstruction priorities.
Prior work on multimodal quantization uses loss-gradient magnitudes to estimate modality sensitivities and incorporates these estimates into the reconstruction objective during calibration~\cite{mbq}. 
We compare this gradient-based sensitivity~\cite{salqvlm, qig} with \emph{commitment discrepancy}, which measures disagreement with FP in selected positions and assigned tokens.
In a pilot experiment on the TextVQA task~\cite{textvqa}, we compute cross-entropy gradients using ground-truth targets at masked answer positions (Fig.~\ref{fig:motivation}(a)) and evaluate commitment changes induced by injecting measured quantization errors into visual or textual conditioning features (Fig.~\ref{fig:motivation}(b)).
In LaViDa-LLaDA-8B~\cite{lavida}, visual features exhibit smaller mean gradient magnitudes yet greater mean commitment discrepancy than textual features.

Motivated by these findings, we propose \textit{CommitQuant}, a post-training quantization framework that aims to preserve FP commitments as closely as possible throughout denoising.
To reflect the partially masked input structure of denoising, \textit{Response-Only Masked Calibration (ROMC)} constructs inputs for both calibration stages by masking only response tokens while keeping the image and prompt visible.
Since visual and textual conditioning errors can affect commitments differently, \textit{Modality-Aware Commitment Calibration (MAC)} uses commitment discrepancy to determine their relative reconstruction priorities.
Block-wise reconstruction can nevertheless leave differences in the final predictions that guide commitment decisions.
To reduce these remaining differences, \textit{Masked-Token KL Refinement (MKR)} aligns the quantized model's prediction distributions with FP at masked positions.

Our analysis in Section~\ref{subsec:discussion} further shows that commitment discrepancy captures quantization-induced changes in denoising decisions, even when loss-gradient magnitudes are small.
Evaluated on LLaDA-V-8B~\cite{lladav}, LaViDa-LLaDA-8B, and LaViDa-Dream-7B~\cite{lavida}, CommitQuant retains up to 99.86\% of BF16 performance under 4-bit weight and activation quantization, with retention averaged across benchmarks for each model. Fig.~\ref{fig:perf_overview} provides per-benchmark comparisons with FP and existing quantization methods on two representative MDLMs.

In summary, our contributions are threefold:
(1) We reveal that loss-gradient sensitivity and commitment discrepancy can yield contrasting modality rankings within the same MDLM, motivating calibration based on the impact of quantization errors on denoising decisions.
(2) We propose CommitQuant, a two-stage post-training quantization framework that combines Modality-Aware Commitment Calibration (MAC) and Masked-Token KL Refinement (MKR), using response-only masked inputs constructed by ROMC.
(3) We evaluate CommitQuant on three MDLMs, demonstrating task performance retention relative to FP and practical generation speedups under W4A4 quantization, with ablations showing improved exact commitment agreement.

\section{Related Work}

\begin{figure*}[t]
\centering
\includegraphics[width=0.99\linewidth]{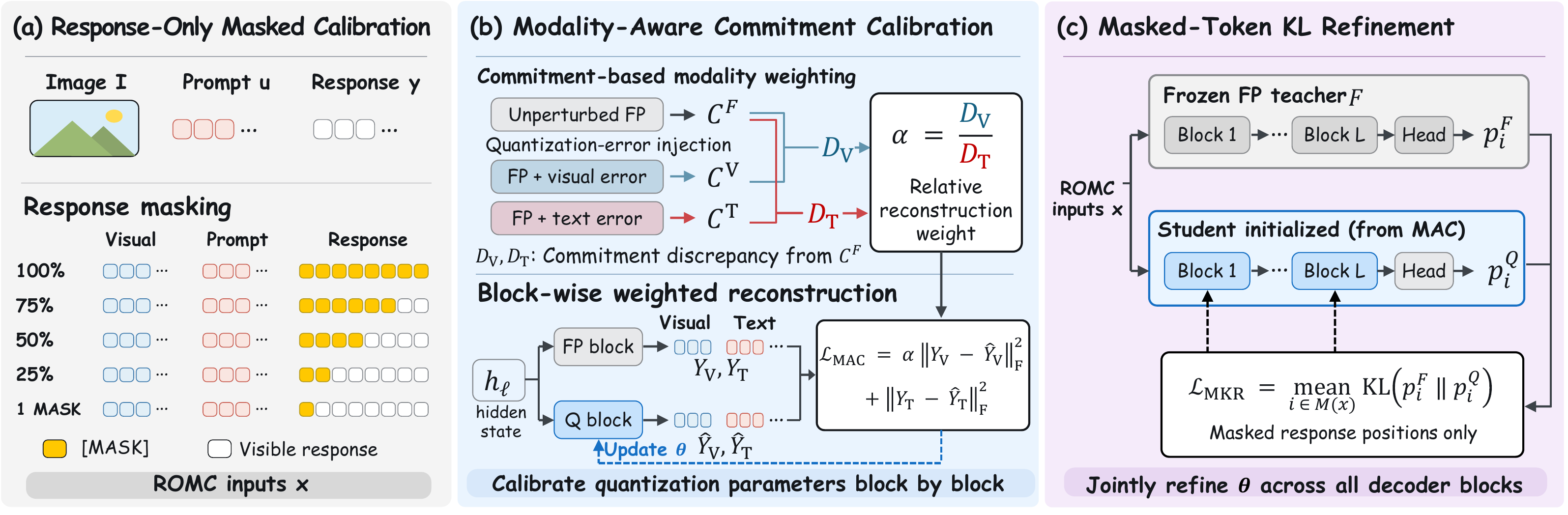}
\vspace{-2mm}
\caption{ \textbf{Overview of CommitQuant.} (a) ROMC constructs calibration inputs at multiple response-masking levels while keeping the image and prompt visible. (b) \textbf{Stage 1: MAC} measures commitment discrepancies by injecting quantization errors separately into visual and textual prefix features on the same FP decoding states. Their ratio, $\alpha = D_V/D_T$, determines the relative visual reconstruction weight for block-wise calibration. (c) \textbf{Stage 2: MKR} starts from the MAC-calibrated model and jointly refines quantization parameters across all decoder blocks by minimizing KL divergence between FP and quantized token distributions at masked response positions under identical inputs. Both stages optimize only quantization parameters while keeping pretrained model weights fixed. }
\vspace{-2mm}
\label{fig:method_overview}
\end{figure*}

\subsection{Multimodal Diffusion Language Models}
\vspace{-1mm}
Masked diffusion language models generate text through iterative masked-token recovery with bidirectional context~\cite{sedd, mdlm, md4}.
LLaDA~\cite{llada} and Dream~\cite{dream} scale this paradigm
through training from scratch and adapting autoregressive
language models, respectively.
LaViDa~\cite{lavida}, LLaDA-V~\cite{lladav}, and Dimple~\cite{dimple} extend it to multimodal understanding, while UniDisc~\cite{unidisc} and MMaDA~\cite{mmada} explore unified diffusion frameworks for multimodal understanding and generation.
Recent efficiency improvements include prefix caching in LaViDa~\cite{lavida} and blockwise decoding with cache reuse in DiffusionVL~\cite{diffusionvl}.
Sparse-LaViDa~\cite{sparselavida} further reduces computation by truncating redundant masked tokens.
Despite these advances, long visual contexts and repeated decoder evaluations remain costly~\cite{lavida, sparselavida}.
We investigate low-bit quantization in this setting, where calibration must account for both multimodal conditioning and changing response states during masked decoding.


\subsection{Quantization} 
\vspace{-1mm}
Post-training quantization (PTQ) reduces model precision for efficient deployment. GPTQ~\cite{gptq} and AWQ~\cite{awq} support weight-only quantization through error compensation and activation-guided scaling, respectively. Weight--activation methods use scaling, clipping, rotations, or learned transformations to mitigate quantization errors~\cite{smoothquant, omniquant, quarot, spinquant, flatquant}.
Modality-aware methods consider differences across token types. MBQ~\cite{mbq} balances visual and textual reconstruction errors using supervised-loss gradients.
ImpQuant~\cite{impquant} estimates token importance from textual attention and visual redundancy.
FreeAct~\cite{freeact} learns separate activation transformations by modality or mask status.

Diffusion-specific methods account for changing decoding states. Quant-dLLM~\cite{quantdllm} samples multiple masking levels, while DLLMQuant~\cite{dllmquant} incorporates temporal, mask, and confidence information. FAIR-Calib~\cite{faircalib} protects fragile commitments using teacher-derived response-position weights. Fast-dVLM~\cite{fastdvlm} integrates FP8 quantization in multimodal block-diffusion inference.
Our method measures changes in selected response positions and tokens under separate visual and textual quantization-error injection.
These measurements reflect the influence of conditioning errors on decoding decisions, distinguishing them from MBQ's loss-gradient weights and FAIR-Calib's response-position priorities.
This connects reconstruction priorities to commitment stability, aiming to preserve visual and textual grounding during low-bit generation.

\section{Methodology}
\label{sec:method}

\paragraph{Notation.}
Let $F$ denote the full-precision (FP) model and $Q_{\theta}$ its quantized counterpart, where $\theta$ denotes the quantization parameters to be calibrated (e.g., transformation matrices, diagonal scaling factors, and weight/activation clipping parameters).
At denoising step $t$, the input state 
$x_t=(I,u,y_t)$ consists of an image $I$, a prompt $u$, and a partially masked response $y_t$,
with masked response positions denoted by $\mathcal{M}(x_t)$.
The decoder predicts tokens and selects masked positions to unmask based on prediction confidence.
Following SPOT~\cite{shlomi2026spot}, an individual \emph{commitment} fixes a predicted token at a selected masked position.
We represent all commitments made in one step as a \emph{commitment set} of position--token pairs, denoted by $C_t^F(x_t)$ and $C_t^{Q_{\theta}}(x_t)$ for the two models.
Applying these commitments replaces the selected \texttt{[MASK]} tokens while leaving all other positions unchanged.
Given the same current state, decoding rule, and schedule, identical commitment sets yield the same next response state, even when prediction distributions differ.
When both models start from the same initial state, equality of commitment sets at every step is therefore sufficient to preserve the FP response-state trajectory.

\paragraph{Overview.}
As illustrated in Fig.~\ref{fig:method_overview},
CommitQuant uses Response-Only Masked Calibration (ROMC) to construct masked calibration inputs by masking response tokens while preserving the image and prompt.
Modality-Aware Commitment Calibration (MAC) derives a relative reconstruction weight from commitment discrepancy and uses it to calibrate $\theta$ block by block.
Starting from MAC, Masked-Token KL Refinement (MKR) jointly refines $\theta$ across all decoder blocks by aligning FP and quantized final prediction distributions at masked response positions under identical inputs.
Both stages use the masked calibration inputs for optimization while keeping pretrained model weights fixed.

\subsection{Response-Only Masked Calibration}
\label{subsec:masked_calibration}
Modality-Aware Commitment Calibration (MAC) and Masked-Token KL Refinement (MKR) use the same masked calibration inputs for transformer-block reconstruction and masked-token prediction alignment, respectively. 
We construct each input $x=(I, u, y)$ by masking response tokens to obtain $y$ while keeping the image $I$ and prompt $u$ unchanged.

Inspired by linear denoising schedules that unmask approximately the same number of tokens per step~\cite{llada, lavida}, we use evenly spaced masking ratios: 100\%, 75\%, 50\%, and 25\% of eligible response tokens. A fifth configuration masks exactly one token to cover near-complete responses. For each example, we randomly permute the eligible response positions once. To mask $k$ tokens, we select the first $k$ entries of this permutation. This yields nested masks that are fixed and reused in both calibration stages. We exclude special and end-of-turn tokens from masking. Section~\ref{subsec:exp_setup} describes the calibration data.


\subsection{Modality-Aware Commitment Calibration}
\label{subsec:MAC}

\begin{table}[t]
    \centering
    \small
    \setlength{\tabcolsep}{4pt}
    \caption{
        Modality weighting for transformer-block reconstruction under W4A4 quantization. Scores are in percentages, with percentage-point changes from the unweighted MSE baseline in parentheses.
    }
        \vspace{-3mm}
    \begin{tabularx}{\columnwidth}
        {@{}>{\raggedright\arraybackslash}Xcc@{}}
        \toprule
        \multirow{2}{*}{Modality weighting}
        & \multicolumn{2}{c}{W4A4} \\
        \cmidrule(lr){2-3}
        & MMMU & AI2D \\
        \midrule
        Baseline
        & 42.89 & 69.53 \\
        \cmidrule[0.25pt](lr){1-3}
        Gradient-based
        & 43.00 {\footnotesize(+0.11)}
        & 69.07 {\footnotesize(-0.46)} \\
        Commitment-based (ours)
        & \textbf{44.00} {\footnotesize(+1.11)}
        & \textbf{69.79} {\footnotesize(+0.26)} \\
        \bottomrule
    \end{tabularx}
    \label{tab:modality_weighting}
    \vspace{-3mm}
\end{table}

Fig.~\ref{fig:motivation} shows contrasting modality rankings on TextVQA.
Text features exhibit larger mean absolute cross-entropy (CE) gradients, while visual quantization errors induce greater mean commitment discrepancy when both perturbations are evaluated on the same FP decoding states.
Using the same TextVQA examples for block-wise reconstruction, we compare commitment-based weighting with CE-gradient-based weighting inspired by MBQ~\cite{mbq}. Without MKR, commitment-based weighting achieves higher scores on both MMMU and AI2D (Table~\ref{tab:modality_weighting}). This result motivates using commitment discrepancy to set the relative reconstruction weight.
The analysis in Section~\ref{subsec:discussion} further motivates this choice by showing that loss-gradient magnitude alone does not provide a general guarantee of commitment stability.

\paragraph{Commitment-based modality weighting.}
To estimate the relative reconstruction weight, we inject quantization errors separately at visual token positions and at textual prefix positions.
We evaluate both perturbations on the same FP decoding states and advance the trajectory using the commitments of $F$.
At each FP decoding state, let $C^F$ denote the commitment set of $F$ and $C^m$ the set obtained under the corresponding perturbation for modality $m\in\{\mathrm{V},\mathrm{T}\}$, where $\mathrm{V}$ and $\mathrm{T}$ denote visual and text tokens, respectively.
We define commitment discrepancy as

\begin{equation}
    D_m
    =
    \mathbb{E}\!\left[
        1 -
        \frac{|C^F \cap C^m|}{|C^F|}
    \right],
    \label{eq:commitment_discrepancy}
\end{equation}
where the expectation denotes the empirical average over
calibration examples and evaluated FP decoding steps.
The intersection requires agreement in both position
and token identity.
The relative reconstruction weight $\alpha = D_{\mathrm{V}}/D_{\mathrm{T}}$ scales visual reconstruction relative to text reconstruction.
We apply the same coefficient across all decoder blocks.


\paragraph{Block-wise weighted reconstruction.}
For each decoder block $\ell$, the FP and quantized blocks receive the same input representations $h_\ell$ obtained from the preceding FP blocks.
Let $Y$ and $\widehat{Y}$ denote their respective outputs over all valid token positions.
Subscripts $\mathrm{V}$ and $\mathrm{T}$ indicate outputs at visual and text token positions, respectively. Using the relative visual weight $\alpha$, we define the reconstruction objective as

\begin{equation}
    \mathcal{L}_{\mathrm{MAC}}
    =
    \alpha
    \left\|Y_{\mathrm{V}}
    - \widehat{Y}_{\mathrm{V}}\right\|_F^2
    +
    \left\|Y_{\mathrm{T}}
    - \widehat{Y}_{\mathrm{T}}\right\|_F^2.
    \label{eq:mac_loss}
\end{equation}

We apply weight $\alpha$ to visual tokens and unit weight to all text tokens. During optimization, we divide the weighted loss by the unweighted reconstruction loss, obtained by setting $\alpha = 1$ in Eq.~\ref{eq:mac_loss}, and stop gradients through the denominator.
We optimize $\theta$ one decoder block at a time using the ROMC inputs. Following FlatQuant~\cite{flatquant}, we optimize the transformation matrices, diagonal scaling factors, and weight and activation clipping parameters while keeping the pretrained model weights fixed.

\subsection{Masked-Token KL Refinement}
\label{subsec:MKR}
Starting from the quantization parameters calibrated by MAC,
MKR aligns the final token predictions used for commitment
decisions.
For each ROMC input $x$, we evaluate $F$ and $Q_{\theta}$
on the identical masked input.
Let $p_i^F(\cdot\mid x)$ and
$p_i^{Q_{\theta}}(\cdot\mid x)$ denote their prediction
distributions over the full vocabulary at position $i$.
We minimize the KL divergence only at masked response
positions:

\begin{equation}
    \mathcal{L}_{\mathrm{MKR}}(x)
    =
    \frac{1}{|\mathcal{M}(x)|}
    \sum_{i\in\mathcal{M}(x)}
    D_{\mathrm{KL}}\!\left(
        p_i^F(\cdot\mid x)
        \,\middle\|\,
        p_i^{Q_{\theta}}(\cdot\mid x)
    \right).
    \label{eq:mkr_loss}
\end{equation}

We average this objective over calibration examples and
jointly refine $\theta$ across all decoder blocks while
keeping pretrained model weights fixed.
This global refinement complements MAC's block-wise
reconstruction by aligning the predictions that guide
token assignment and confidence-based position selection,
thereby encouraging agreement with FP commitments.

\subsection{Discussion}
\label{subsec:discussion}
A main observation motivating CommitQuant is that visual-token features exhibit smaller mean loss-gradient magnitudes than text-token features, while visual-token quantization errors induce greater mean commitment discrepancy (Fig.~\ref{fig:motivation}). The following analysis further explains \textit{why a smaller loss gradient need not imply higher commitment stability: a perturbation can change a discrete decision while leaving the loss unchanged to first order}.

For a given partially masked input $x=(I,u,y)$, $h$ denotes the vectorized FP input to decoder block $\ell$, and $\mathcal{L}(h)$ is the masked-token cross-entropy loss. The downstream logit of vocabulary token $v\in \mathcal{V}$ at masked response position $i\in\mathcal{M}(x)$ is denoted by $s_{i,v}(h)$. Under greedy token prediction, the confidence at position $i$ is the softmax probability of the predicted token: 
\begin{equation} 
c_i(h) = \max_{v \in \mathcal{V}} \frac{\exp(s_{i,v}(h))} {\sum_{v' \in \mathcal{V}} \exp(s_{i,v'}(h))}. 
\end{equation}
Confidence-based top-$k$ decoding selects the $k$ masked response positions with the largest $c_i(h)$.

For a fixed $k$, we rank the masked response positions by their FP confidence in descending order. The positions ranked $k$-th and $(k+1)$-th are denoted by $i$ and $j$, respectively.
Assuming no tie at the selection cutoff, their selection margin is defined as
\begin{equation}
    m(h) = c_i(h) - c_j(h) > 0.
\end{equation}
To examine the stability of this ranking, we perturb $h$ and compare the confidence values at the same positions
$i$ and $j$ before and after perturbation.
If the margin becomes negative, the originally unselected position $j$ outranks the selected position $i$.
The FP position-selection set therefore cannot be preserved, changing the commitments.
\vspace{-3mm}
\paragraph{Do small loss gradients guarantee stable commitments?} 
Assume the logits are twice continuously differentiable
in a neighborhood of $h$ where predicted token identities
remain unchanged.
If $\nabla_h m(h)$ has a nonzero component orthogonal to
$\nabla_h\mathcal{L}(h)$, there exists a unit perturbation
direction $d$ satisfying
\begin{equation}
    \bigl(\nabla_h\mathcal{L}(h)\bigr)^\top d = 0,
    \qquad
    \bigl(\nabla_h m(h)\bigr)^\top d < 0.
\end{equation}

Since the loss has zero directional derivative along $d$,
Taylor's theorem gives, for sufficiently small $\varepsilon>0$,
\begin{equation}
    \left|\mathcal{L}(h+\varepsilon d)-\mathcal{L}(h)\right|
    \leq \tfrac{1}{2}B_L\varepsilon^2,
    \label{eq:loss_change_bound}
\end{equation}
where $B_L$ bounds the Hessian operator norm of $\mathcal{L}$ in this neighborhood.
For the selection margin, Taylor's theorem gives
\begin{equation}
    m(h+\varepsilon d)
    \leq m(h)
    + \varepsilon\bigl(\nabla_h m(h)\bigr)^\top d
    + \tfrac{1}{2}B_m\varepsilon^2,
    \label{eq:margin_change_bound}
\end{equation}
where $B_m$ bounds the Hessian operator norm of
$m$ in the same neighborhood. If the upper bound in Eq.~\ref{eq:margin_change_bound}
is negative for such an $\varepsilon$ within this neighborhood, the FP position-selection set and commitments change, while the loss change remains bounded as in Eq.~\ref{eq:loss_change_bound}.
Thus, commitments can change even when the first-order change in CE is zero. Section~\ref{sec:analysis} shows empirically that measured quantization errors can change commitments even when the resulting change in CE is small.

\begin{tcolorbox}[
    colback=gray!8,
    colframe=gray!50,
    boxrule=0.4pt,
    sharp corners,
    left=5pt,
    right=5pt,
    top=4pt,
    bottom=4pt
]
\textbf{Takeaway.}
Loss gradients alone may overlook changes in denoising
decisions that can affect generation quality, motivating
commitment discrepancy as a direct measure of
quantization sensitivity.
\end{tcolorbox}

\section{Experiments}


\begin{table*}[t!]
\scriptsize
\centering
\caption{Performance comparison on LLaDA-V-8B and LaViDa-LLaDA-8B
across seven multimodal benchmarks, including the Prefix-DLM
configuration of LaViDa.
Quantized models use W4A4KV16.
Avg.\ denotes the arithmetic mean of per-benchmark score ratios
to the corresponding full-precision model, expressed as a percentage.
Bold indicates the best reported result among quantized methods.}
\label{tab:mdllm-seven-benchmarks}
\vspace{-1mm}

\resizebox{0.96\textwidth}{!}{%
\setlength{\tabcolsep}{7pt}
\begin{tabular}{l|ccccccc|c}
\toprule
\textbf{Method} &
\shortstack{\textbf{MMMU} \\ Acc. $\uparrow$} &
\shortstack{\textbf{MME} \\ P+C $\uparrow$} &
\shortstack{\textbf{AI2D} \\ Acc. $\uparrow$} &
\shortstack{\textbf{MMStar} \\ Avg. $\uparrow$} &
\shortstack{\textbf{POPE} \\ F1 $\uparrow$} &
\shortstack{\textbf{TextVQA} \\ VQA Score $\uparrow$} &
\shortstack{\textbf{SEED-I} \\ Acc. $\uparrow$} &
\shortstack{\textbf{Avg.} \\ Rel. (\%) $\uparrow$} \\
\midrule

\rowcolor{gray!20}
\multicolumn{9}{c}{\textit{LLaDA-V-8B}} \\
\midrule
Original (FP16)
    & 49.00 & 1994.99 & 77.78 & 60.37
    & 86.86 & 64.36 & 74.78 & 100.00 \\
\midrule
QuaRot {\textcolor{gray}{(NeurIPS 2024)}}
    & 47.44 & \textbf{1975.48} & 77.30 & 57.25
    & 86.60 & 60.27 & 74.17 & 97.51 \\
SpinQuant {\textcolor{gray}{(ICLR 2025)}}
    & 45.67 & 1870.63 & 75.84 & 57.34
    & 86.49 & 59.38 & 74.21 & 95.79 \\
FlatQuant {\textcolor{gray}{(ICML 2025)}}
    & 46.78 & 1965.83 & 77.10 & 58.87
    & 86.70 & \textbf{63.38} & 74.43 & 98.35 \\
\rowcolor[RGB]{232,225,245}
CommitQuant (ours)
    & \textbf{48.11} & 1958.55 & \textbf{77.33} & \textbf{59.41}
    & \textbf{86.98} & 63.07 & \textbf{74.63}
    & {\textbf{98.87}} \\
\midrule

\rowcolor{gray!20}
\multicolumn{9}{c}{\textit{LaViDa-LLaDA-8B (Full-DLM)}} \\
\midrule
Original (BF16)
    & 44.33 & 1722.44 & 69.27 & 45.72
    & 83.41 & 58.84 & 66.42 & 100.00 \\
\midrule
QuaRot {\textcolor{gray}{(NeurIPS 2024)}}
    & 42.22 & 1682.27 & 68.56 & 46.16
    & 82.94 & 57.78 & 66.04 & 98.56 \\
SpinQuant {\textcolor{gray}{(ICLR 2025)}}
    & 42.11 & 1642.67 & 68.36 & 46.20
    & 82.65 & 57.92 & 65.39 & 98.01 \\
FlatQuant {\textcolor{gray}{(ICML 2025)}}
    & 43.11 & 1624.46 & 69.13 & 46.11
    & 83.13 & \textbf{58.38} & 66.30 & 98.70 \\
\rowcolor[RGB]{232,225,245}
CommitQuant (ours)
    & \textbf{44.22} & \textbf{1686.55} & \textbf{69.27} & \textbf{46.34}
    & \textbf{83.53} & 58.27 & \textbf{66.98}
    & {\textbf{99.86}} \\
\midrule

\rowcolor{gray!20}
\multicolumn{9}{c}{\textit{LaViDa-LLaDA-8B (Prefix-DLM)}} \\
\midrule
Original (BF16)
    & 44.22 & 1694.40 & 70.08 & 46.87
    & 83.75 & 56.33 & 66.61 & 100.00 \\
\midrule
QuaRot {\textcolor{gray}{(NeurIPS 2024)}}
    & 41.89 & 1691.96 & 67.94 & 45.46
    & 83.36 & 54.80 & 66.89 & 97.97 \\
SpinQuant {\textcolor{gray}{(ICLR 2025)}}
    & 41.00 & 1688.22 & 67.26 & 43.75
    & 83.62 & 54.50 & 65.42 & 96.64 \\
FlatQuant {\textcolor{gray}{(ICML 2025)}}
    & 42.78 & 1641.03 & 68.78 & \textbf{45.52}
    & \textbf{84.13} & 55.42 & 66.54 & 98.23 \\
\rowcolor[RGB]{232,225,245}
CommitQuant (ours)
    & \textbf{43.22} & \textbf{1702.46} & \textbf{69.14} & 45.41
    & 84.00 & \textbf{55.61} & \textbf{66.92}
    & {\textbf{99.03}} \\
\bottomrule
\end{tabular}%
}
\vspace{-1mm}
\end{table*}

\begin{table*}[t!]
\scriptsize
\centering
\caption{Performance comparison on LaViDa-Dream-7B across seven
multimodal benchmarks under Full-DLM and Prefix-DLM.}
\label{tab:dream-eight-benchmarks}
\vspace{-3mm}

\resizebox{0.95\textwidth}{!}{%
\setlength{\tabcolsep}{7pt}
\begin{tabular}{l|ccccccc|c}
\toprule
Method &
\shortstack{MMMU \\ Acc. $\uparrow$} &
\shortstack{MME \\ P+C $\uparrow$} &
\shortstack{AI2D \\ Acc. $\uparrow$} &
\shortstack{MMStar \\ Avg. $\uparrow$} &
\shortstack{POPE \\ F1 $\uparrow$} &
\shortstack{TextVQA \\ VQA Score $\uparrow$} &
\shortstack{SEED-I \\ Acc. $\uparrow$} &
\shortstack{Avg. \\ Rel. (\%) $\uparrow$} \\
\midrule

\rowcolor{gray!20}
\multicolumn{9}{c}{\textit{LaViDa-Dream-7B (Full-DLM)}} \\
\midrule
Original (BF16)
    & 42.33 & 1839.15 & 68.43 & 46.21
    & 85.91 & 57.45 & 69.75 & 100.00 \\
\midrule
QuaRot {\textcolor{gray}{(NeurIPS 2024)}}
    & 40.33 & 1805.76 & 67.68 & \textbf{45.96}
    & 85.12 & 56.38 & 68.79 & 98.24 \\
SpinQuant {\textcolor{gray}{(ICLR 2025)}}
    & 40.78 & 1778.91 & 68.01 & 44.52
    & 85.13 & 54.99 & 68.76 & 97.45 \\
FlatQuant {\textcolor{gray}{(ICML 2025)}}
    & 40.67 & 1768.59 & \textbf{68.48} & 45.91
    & \textbf{86.20} & 56.32 & 69.26 & 98.48 \\
\rowcolor[RGB]{232,225,245}
CommitQuant (ours)
    & \textbf{41.56} & \textbf{1838.74} & 67.81 & 45.19
    & 85.77 & \textbf{56.71} & \textbf{69.39} & \textbf{99.01} \\
\midrule

\rowcolor{gray!20}
\multicolumn{9}{c}{\textit{LaViDa-Dream-7B (Prefix-DLM)}} \\
\midrule
Original (BF16)
    & 42.33 & 1856.16 & 68.85 & 45.30
    & 85.84 & 57.03 & 69.73 & 100.00 \\
\midrule
QuaRot {\textcolor{gray}{(NeurIPS 2024)}}
    & 39.78 & 1756.27 & 67.00 & 43.55
    & 85.44 & 56.31 & 68.85 & 97.01 \\
SpinQuant {\textcolor{gray}{(ICLR 2025)}}
    & 40.56 & 1747.62 & 66.68 & 43.45
    & 84.96 & 54.39 & 68.62 & 96.50 \\
FlatQuant {\textcolor{gray}{(ICML 2025)}}
    & 42.11 & 1754.16 & \textbf{68.43} & 44.60
    & \textbf{86.20} & 56.17 & 69.37 & 98.60 \\
\rowcolor[RGB]{232,225,245}
CommitQuant (ours)
    & \textbf{42.22} & \textbf{1817.04} & 68.20 & \textbf{44.81}
    & 85.73 & \textbf{56.46} & \textbf{69.44} & \textbf{99.15} \\
\bottomrule
\end{tabular}%
}
    \vspace{-3mm}
\end{table*}

\subsection{Experimental Setup}
\label{subsec:exp_setup}
\paragraph{Models and datasets.}
We evaluate CommitQuant on LLaDA-V-8B~\cite{lladav}, LaViDa-LLaDA-8B, and LaViDa-Dream-7B~\cite{lavida}.
We compare against the corresponding full-precision (FP) models and three post-training quantization methods: QuaRot~\cite{quarot}, SpinQuant~\cite{spinquant}, and FlatQuant~\cite{flatquant}. We also evaluate MBQ~\cite{mbq} and ImpQuant~\cite{impquant} and report their results in the appendix.
We evaluate performance across seven multimodal benchmarks: MMMU~\cite{mmmu}, MME~\cite{mme}, AI2D~\cite{ai2d}, MMStar~\cite{mmstar}, POPE~\cite{pope}, TextVQA~\cite{textvqa}, and SEED-I~\cite{seedbench}. All methods follow identical evaluation protocols within each model configuration. For LLaDA-V, we follow its official task-specific generation settings. We detail the evaluation and decoding settings in the appendix.

\vspace{-2mm}
\paragraph{Quantization and calibration.}
For benchmark accuracy evaluation, all quantized methods use W4A4KV16 with per-channel weight quantization and per-token activation quantization.
We calibrate all quantized methods on the same 128 image--caption pairs from the COCO 2017 training split~\cite{coco}, using \texttt{Describe the image.} as the prompt for each image. For ROMC, we mask selected tokens in each human-annotated COCO caption to obtain the masked response $y$.
Most experiments are conducted using four NVIDIA GH200 GPUs, while the inference efficiency experiments in Section~\ref{sec:inference_efficiency} are performed on a single NVIDIA A100 GPU.
We list the optimization hyperparameters for MAC and MKR in the appendix.

\vspace{-2mm}

\paragraph{Inference configurations.}
Following LaViDa~\cite{lavida}, we evaluate both LaViDa models under Full-DLM with full attention and Prefix-DLM with prefix KV caching. In Prefix-DLM, image and prompt tokens attend only within the prefix, while response tokens attend to the entire sequence. For each model and quantization method, we use the same calibrated quantization parameters and keep all other evaluation settings fixed across the two configurations.

\vspace{-1mm}


\subsection{Comparison on Diverse Tasks}

\paragraph{LLaDA-based models.}
In Table~\ref{tab:mdllm-seven-benchmarks}, CommitQuant achieves the highest average relative performance among quantized methods, retaining 98.87\% of FP performance on LLaDA-V-8B and 99.86\% and 99.03\% on LaViDa-LLaDA-8B under Full-DLM and Prefix-DLM, respectively.
Compared with FlatQuant, CommitQuant raises MMMU accuracy on LLaDA-V-8B from 46.78\% to 48.11\% and improves MME by over 61 points under both LaViDa-LLaDA-8B configurations.

\vspace{-3mm}
\paragraph{Dream-based models.}
To test whether CommitQuant extends beyond LLaDA~\cite{llada}, which was trained from scratch with masked diffusion, we evaluate LaViDa-Dream-7B, whose backbone adapts autoregressive Qwen2.5-7B~\cite{qwen25} weights to diffusion while retaining shifted token prediction~\cite{dream}. In Table~\ref{tab:dream-eight-benchmarks}, CommitQuant again leads among quantized methods, retaining 99.01\% and 99.15\% of FP performance under Full-DLM and Prefix-DLM, respectively.


\begin{table}[t]
    \centering
    \caption{
        Ablation of MAC and MKR on MMMU and MME.
        All configurations use response-only masked calibration (ROMC).
        The first row reports the ROMC-only baseline.
    }
    \label{tab:calibration_loss_ablation}
    \footnotesize
    \setlength{\tabcolsep}{3.5pt}
    \resizebox{\columnwidth}{!}{%
    \begin{tabular}{@{}cc|c|ccc@{}}
        \toprule
        \multicolumn{2}{c|}{\textbf{Calibration components}} &
        \textbf{MMMU} & \multicolumn{3}{c}{\textbf{MME}} \\
        \makebox[5em][c]{MAC} & \makebox[5em][c]{MKR} &
        Acc. (\%) $\uparrow$ &
        Perception $\uparrow$ &
        Cognition $\uparrow$ &
        Total $\uparrow$ \\
        \midrule
                     &              &
        41.33 & 1282.65 & 321.07 & 1603.73 \\
        $\checkmark$ &              &
        43.11 & 1319.23 & 303.93 & 1623.16 \\
                     & $\checkmark$ &
        44.00 & 1303.57 & 303.21 & 1606.78 \\
        $\checkmark$ & $\checkmark$ &
        \textbf{44.22} & \textbf{1322.26} &
        \textbf{364.29} & \textbf{1686.55} \\
        \bottomrule
    \end{tabular}%
    }
    \vspace{-2mm}
\end{table}

\begin{figure}[t]
    \centering
    \hspace{-2mm}
    \includegraphics[width=1.03\columnwidth]{
        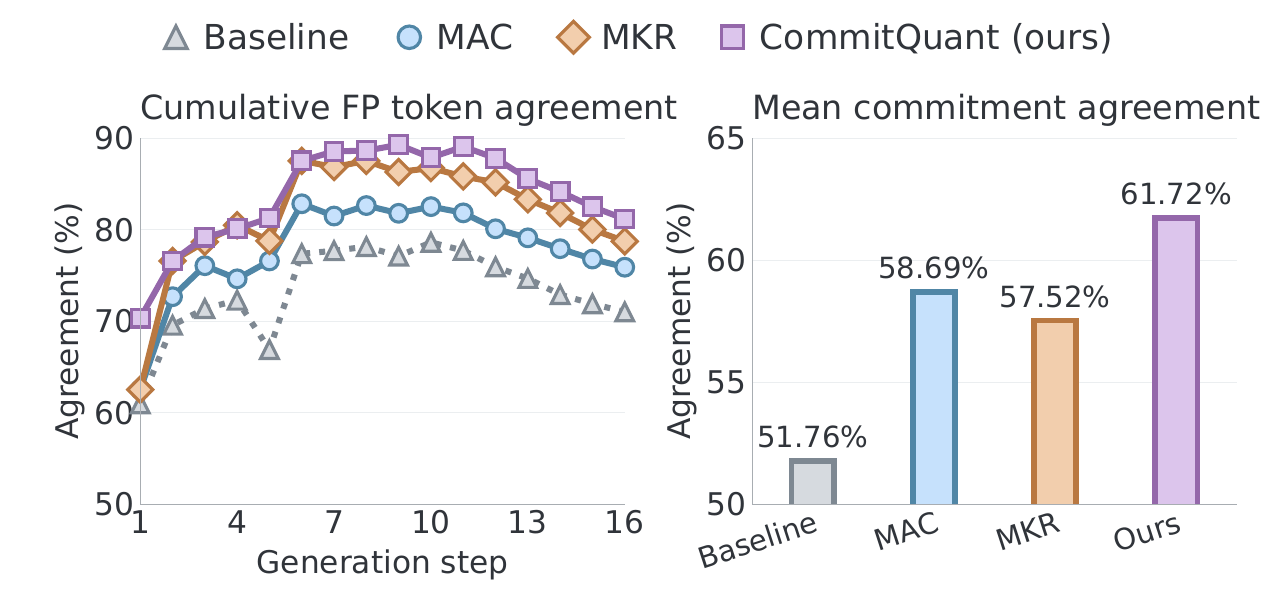
    }
    \vspace{-4mm}
    \caption{
        \textbf{Commitment agreement ablation.}
        (Left) Cumulative token agreement at FP-committed positions.
        (Right) Mean exact agreement with FP on newly committed position--token pairs over 64 COCO images and 16 generation steps.
        Baseline denotes ROMC alone.
    }
        \vspace{-4mm}
    \label{fig:commitment_ablation}
\end{figure}

\subsection{Ablation Study}

\begin{figure}[t]
    \centering
    \includegraphics[
        width=\columnwidth
    ]{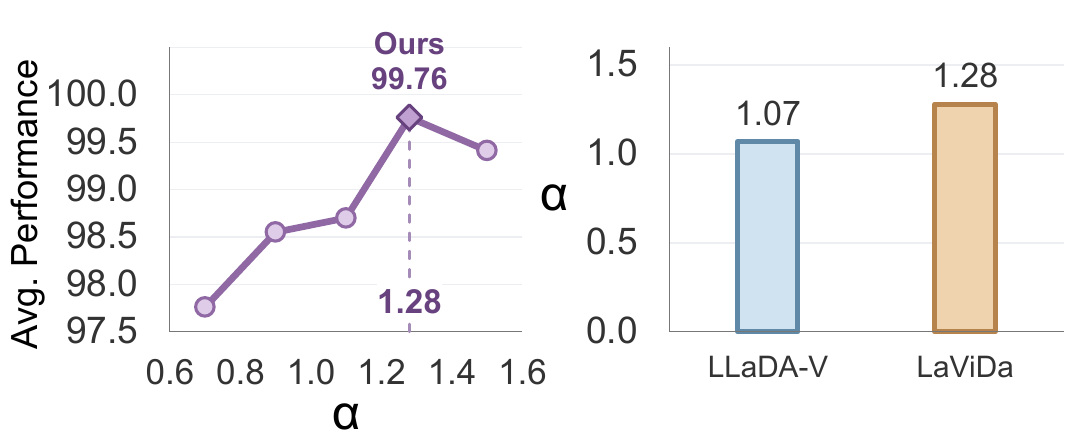}
    \vspace{-6mm}
    \caption{
        \textbf{Analysis of the visual reconstruction weight $\alpha$.}
        (Left) Average performance on LaViDa across tested values of $\alpha$,
        with the commitment-based choice ($\alpha=1.28$) performing best.
        (Right) Commitment-based weights for LLaDA-V and LaViDa.
    }
    \label{fig:alpha_analysis}
    \vspace{-1mm}
\end{figure}

\paragraph{Effect of calibration components.}
Table~\ref{tab:calibration_loss_ablation} evaluates MAC and MKR against an unweighted block-wise reconstruction baseline ($\alpha=1$), using the same response-masked inputs across configurations (Section~\ref{subsec:masked_calibration}). MAC and MKR each improve MMMU and total MME over the baseline. Their combination yields the best results, reaching 44.22\% on MMMU and 1686.55 on MME. Adding MKR after MAC raises total MME by 63.39 points, compared with a 3.05-point gain when adding MKR directly to the baseline.

\begin{figure}[t]
    \centering
    \includegraphics[
        width=\columnwidth
    ]{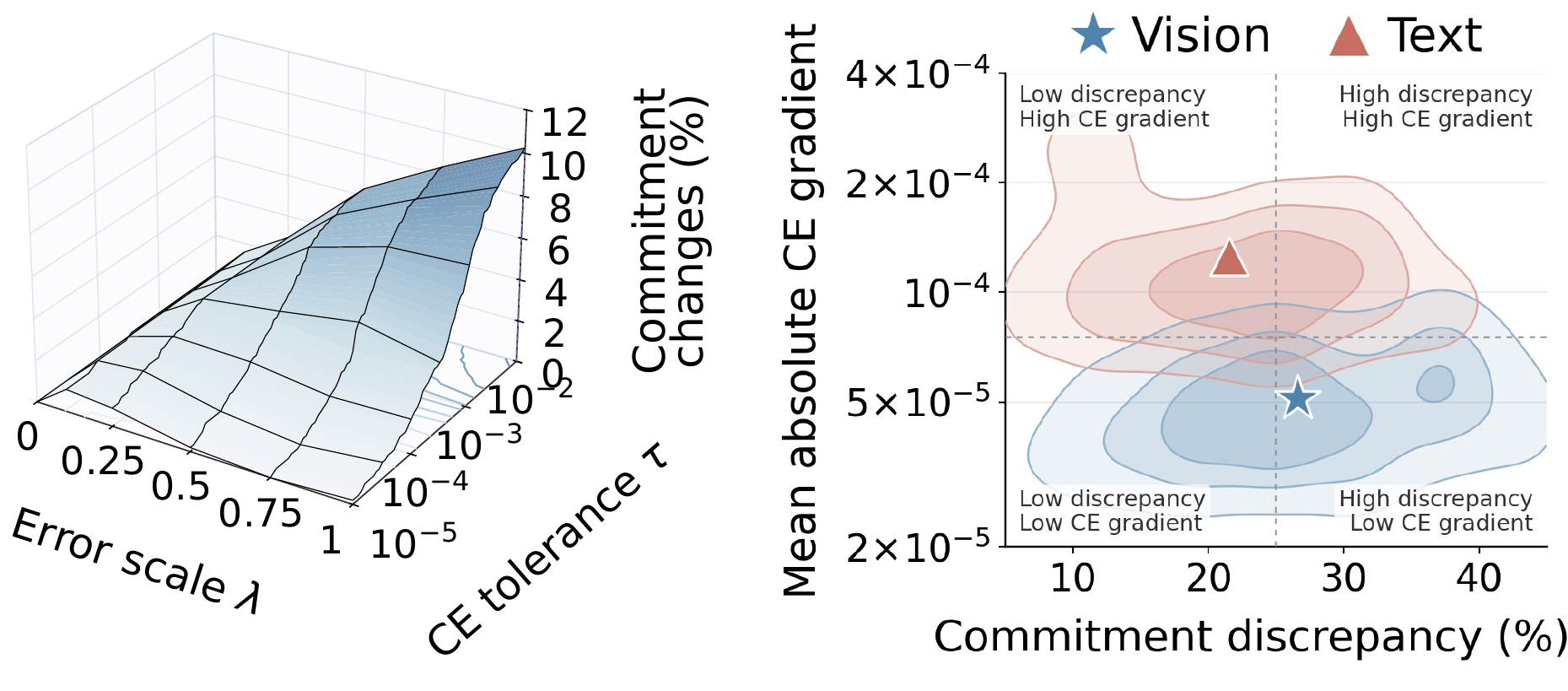}

    \caption{
        \textbf{CE sensitivity and commitment stability.}
        (Left) Commitment change rate under CE-gradient-orthogonal quantization errors, varying error scale $\lambda$ and CE tolerance $\tau$.
        (Right) Mean absolute CE gradient versus commitment discrepancy.
        Markers denote modality means, and dashed lines mark pooled medians.
    }
    \label{fig:gradeint_analysis}
    \vspace{-2mm}
\end{figure}

\begin{figure*}[t]
\centering
\includegraphics[width=1.0\linewidth]{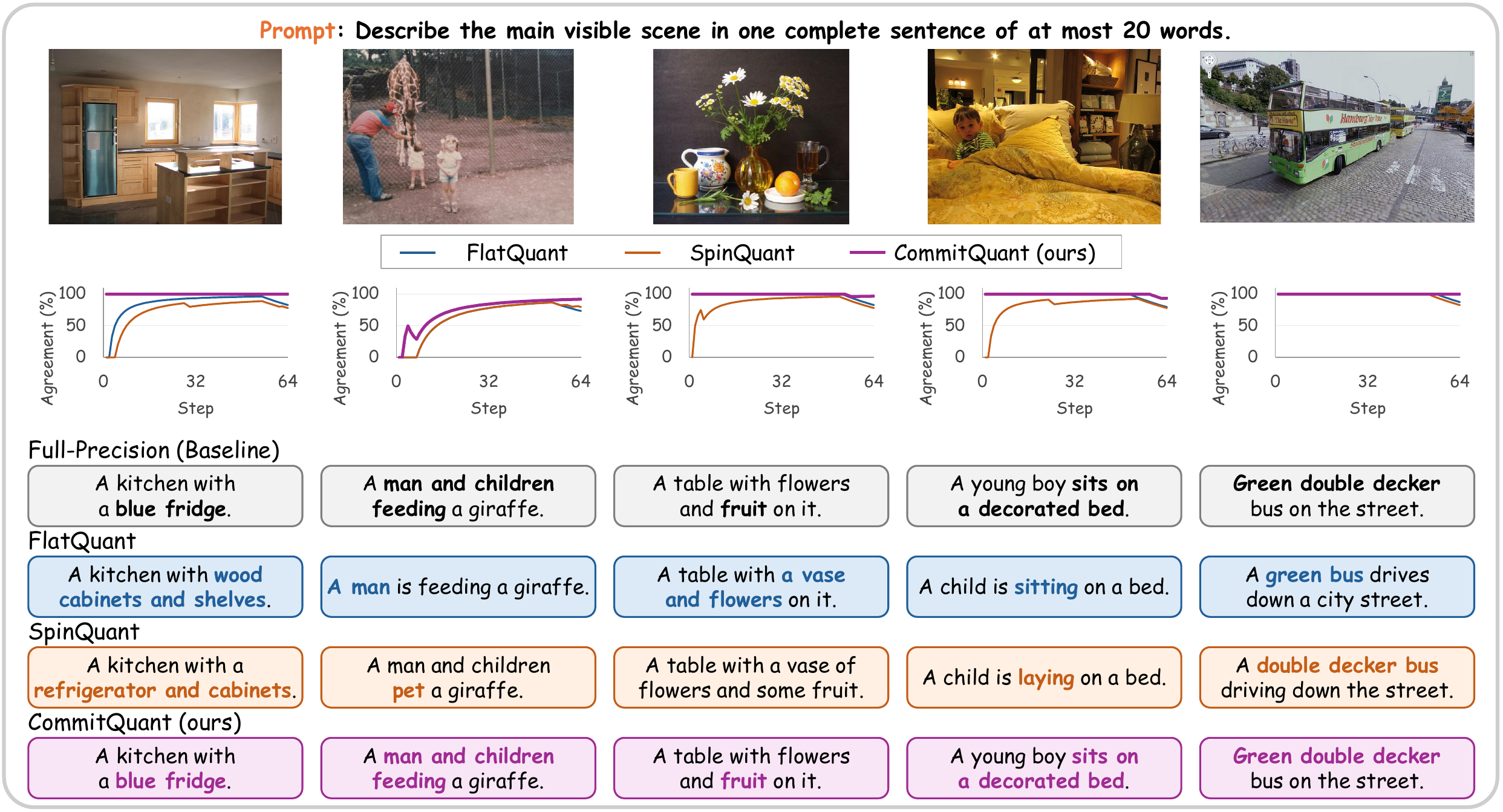}
\vspace{-3mm}
\caption{ \textbf{Qualitative comparison with full precision.} Curves show cumulative same-step position-and-token agreement with FP over 64 decoding steps on five selected COCO images, including end-of-turn and trailing end-of-text tokens. CommitQuant matches all five FP descriptions and achieves the highest final agreement among quantized methods in each example.}
\label{fig:case_study}
\vspace{-3mm}
\end{figure*}

\vspace{-3mm}
\paragraph{Commitment agreement with full precision.}
Fig.~\ref{fig:commitment_ablation} examines whether the task-level gains are accompanied by closer agreement with full precision (FP) during generation.
Mean exact agreement on newly committed position--token pairs increases from 51.76\% for the baseline to 58.69\% with MAC and 61.72\% after adding MKR, a total gain of 9.96 percentage points. 
CommitQuant also achieves higher cumulative token agreement at FP-committed positions across decoding steps.

\vspace{-3mm}
\paragraph{Effect of the Relative Reconstruction Weight.}
The left panel of Fig.~\ref{fig:alpha_analysis} compares four fixed weights ($\alpha \in \{0.7, 0.9, 1.1, 1.5\}$) with the weight derived from the ratio of visual to text commitment discrepancies ($\alpha=1.28$) on LaViDa-LLaDA. This derived weight achieves the highest average score among all evaluated settings (99.76), compared with 99.41 at $\alpha=1.5$. The same rule yields weights of 1.07 for LLaDA-V and 1.28 for LaViDa, emphasizing visual reconstruction in both models to different degrees.

\subsection{Analysis}
\label{sec:analysis}

\paragraph{Loss Gradients and Commitment Stability.}
Fig.~\ref{fig:gradeint_analysis} provides empirical evidence for the analysis in Section~\ref{subsec:discussion}. In the left panel, we project measured quantization errors onto the subspace orthogonal to the CE gradient and inject the scaled projection at the output of one decoder block at a time, across all visual and textual token positions. The projection sets the first-order effect on CE to zero.
For each error scale and tolerance $\tau$, we report the fraction of 1,024 COCO images with at least one block intervention that changes a commitment while satisfying $|\Delta\mathrm{CE}|\leq\tau$ and a relative CE change of at most 1\%. At unit error scale and $\tau=10^{-2}$, 105 images meet these criteria.

The right panel shows a corresponding mismatch across modalities. Over 128 COCO images, visual features exhibit smaller mean absolute CE gradients than textual features, yet visual perturbations induce greater mean commitment discrepancy. Because each commitment determines which positions and tokens enter the next denoising state, these changes can affect the final response. Thus CE-gradient magnitude alone can misrank modalities by their sensitivity to quantization during generation. This motivates using commitment discrepancy to set the relative reconstruction weight.

\begin{figure}[t]
    \centering
    \hspace{-2mm}
    \includegraphics[width=\columnwidth]{
        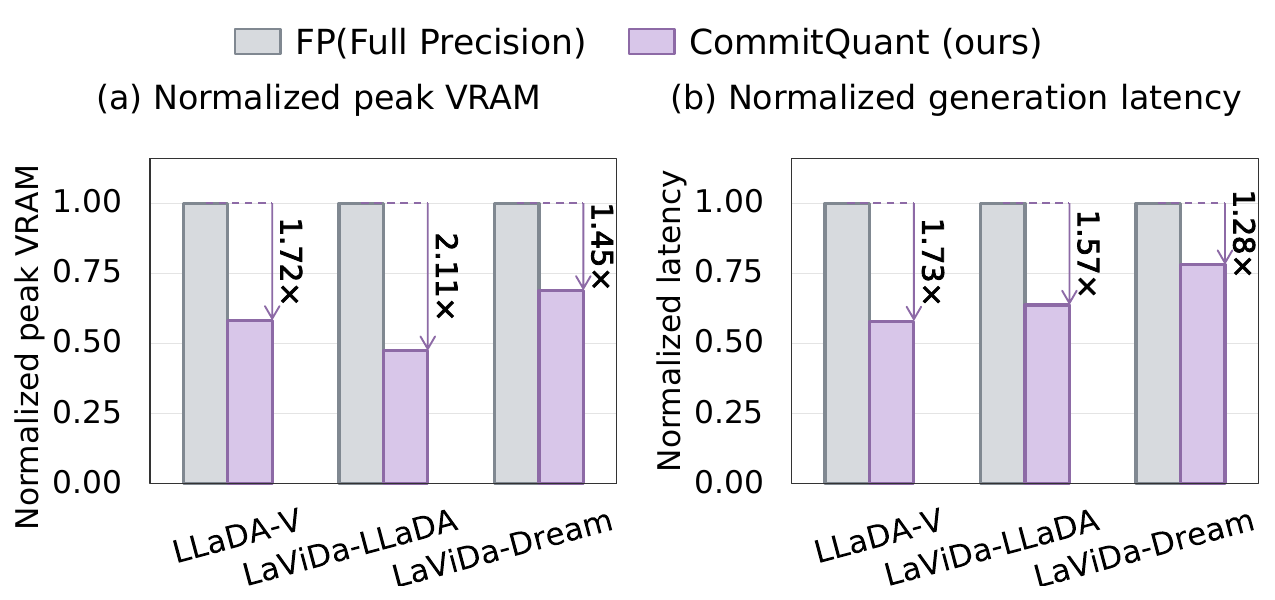
    }
    \vspace{-3mm}
    \caption{
    \textbf{Efficiency of CommitQuant.}
        (a) Peak inference VRAM and (b) generation latency of CommitQuant,
        normalized to each model's full-precision (FP) baseline.
        Arrows indicate memory reduction ratios and speedups over FP.
    }
    \label{fig:inference_efficiency}
        \vspace{-3mm}
\end{figure}

\vspace{-3mm}
\paragraph{Qualitative Comparison.}
Fig.~\ref{fig:case_study} compares FP, FlatQuant, SpinQuant, and CommitQuant on five selected COCO images.
FlatQuant and SpinQuant omit or alter details such as the refrigerator's color and the child's posture.
CommitQuant reproduces all five FP responses and exceeds the better of the two quantized baselines by 12.50--17.19 percentage points in final cumulative same-step position-and-token agreement.
\vspace{-3mm}
\paragraph{Latency and Memory Usage.}
\label{sec:inference_efficiency}
We evaluate generation latency and peak inference VRAM across three MDLMs on a single NVIDIA A100 GPU.
Fig.~\ref{fig:inference_efficiency} compares CommitQuant with the corresponding full-precision (FP) baselines, with both metrics normalized to FP.
CommitQuant achieves 1.28--1.73$\times$ speedups and reduces peak inference VRAM by 31.2--52.7\% across the evaluated models. These results demonstrate that CommitQuant enables faster generation with lower GPU memory requirements. 
We adapt FlatQuant's deployment design~\cite{flatquant} for W4A4 inference on the A100.

\vspace{-2mm}
\section{Conclusion}
CommitQuant targets FP commitment preservation for low-bit MDLM quantization, motivated by the contrasting modality rankings under loss-gradient and commitment-based sensitivity measures. It combines modality-aware block-wise reconstruction with masked-token KL refinement using response-only masked calibration inputs. Experiments on LaViDa and LLaDA-V show improved overall multimodal performance under W4A4 quantization, while ablations demonstrate greater agreement with FP commitments. However, the current framework uses a single visual reconstruction weight, which does not explicitly capture variations in modality sensitivity across decoder blocks and denoising steps. Future work will explore block- and step-adaptive weighting to address this limitation.

\section*{Acknowledgments}
This work was supported by Korea University Grants (Nos. K2631131, K2630931, and  K2631761).
{
    \small
    \bibliographystyle{ieeenat_fullname}
    \bibliography{main}
}

\clearpage
\maketitlesupplementary
\appendix
\etocdepthtag.toc{supplement}



\begingroup
\small
\setcounter{tocdepth}{2}
\begingroup
    \etocsettagdepth{main}{none}
    \etocsettagdepth{supplement}{subsection}
    \etocsetnexttocdepth{subsection}
    \tableofcontents
\endgroup
\medskip
\endgroup

\section{Additional Implementation and Evaluation Details}
\label{sec:supp-implementation}

\subsection{Models and Quantization Settings}
\label{sec:supp-quantization}

We evaluate CommitQuant on the language backbones of
LaViDa-LLaDA-8B and LaViDa-Dream-7B~\cite{lavida},
and LLaDA-V-8B~\cite{lladav} under W4A4KV16: weights and activations are quantized
to 4 bits, while keys and values retain 16-bit precision.
We use symmetric per-channel weight quantization and
dynamic symmetric per-token activation quantization.
The vision encoder and multimodal projector retain
their original precision.

We adopt the FlatQuant~\cite{flatquant} parameterization and optimize
transformation matrices, diagonal scales, and clipping
parameters while keeping pretrained model weights frozen.
Accuracy evaluation uses fake quantization with fused
transformations to simulate the specified quantization
setting.

\subsection{Calibration Data}
\label{sec:supp-calibration}

We use 128 image--caption pairs selected from the COCO2017
training split~\cite{coco}. Each input consists of an image,
the prompt \texttt{Describe the image.}, and its
human-annotated reference caption, formatted using the
model's native template. The caption supplies the response
tokens used to construct ROMC inputs; the image and prompt
remain visible in every masking state.

For a caption containing $L$ body tokens, we construct
five nested masking states with $L$,
$\lceil 0.75L \rceil$, $\lceil 0.5L \rceil$,
$\lceil 0.25L \rceil$, and one masked token, respectively.
Masked positions are selected from a fixed token permutation
for each example, while special tokens remain unchanged.
Each image contributes one state per epoch, cycling through
all five states over five epochs.
MAC and MKR use the same calibration pairs and masking
schedule.

\subsection{Commitment Discrepancy Measurement}
\label{sec:supp-discrepancy}

We measure commitment discrepancy along the native FP
generation trajectory, without providing reference captions.
The response is initially fully masked; subsequent states
contain previously committed FP tokens and the remaining masks.
At each step, all interventions use the same FP state.
Independent FP and MSE-calibrated quantized forward passes
on this state provide block-input representations.
At each decoder block, we add their Q-minus-FP difference
to either the visual token positions ($\mathrm{V}$) or
the textual prefix positions ($\mathrm{T}$) in a separate
FP forward pass.
The perturbations accumulate across blocks, and the resulting
logits determine the commitment.
Each intervention is evaluated independently, and subsequent
states always follow FP commitments.

A commitment consists of the selected response position
and token ID.
We record a discrepancy if either component differs
from the FP decision:

\begin{equation}
    D_s =
    \frac{1}{NT}
    \sum_{i=1}^{N}\sum_{t=1}^{T}
    \mathbf{1}\!\left[
    a_s(x^{\mathrm{FP}}_{i,t})
    \neq
    a_{\mathrm{FP}}(x^{\mathrm{FP}}_{i,t})
    \right],
    \quad s\in\{\mathrm{V},\mathrm{T}\},
\end{equation}

where $N$ is the number of images and $T$ is the number
of generation steps.
The modality ratio $\alpha=D_{\mathrm{V}}/D_{\mathrm{T}}$ is
computed from aggregate discrepancy counts rather than
averaging per-image ratios.
A zero text discrepancy yields an undefined ratio.

\subsection{MAC and MKR Optimization}
\label{sec:supp-optimization}

Algorithms~\ref{alg:mac} and~\ref{alg:mkr} summarize block-wise MAC and joint MKR. For block $\ell$, $F_\ell$ and $Q_{\theta_\ell}$ denote the FP and quantized blocks, with quantization parameters $\theta_\ell$.

\begin{algorithm}[t]
\caption{MAC optimization for decoder block $\ell$}
\label{alg:mac}
\begin{algorithmic}[1]
\Require FP model $F$, quantized block $Q_{\theta_\ell}$, ROMC inputs $\mathcal{D}$, relative reconstruction weight $\alpha=D_{\mathrm{V}}/D_{\mathrm{T}}$
\For{each optimization iteration}
    \State Sample a minibatch from $\mathcal{D}$
    \State Obtain $h_\ell$ from the preceding FP blocks
    \State $Y \gets F_\ell(h_\ell)$
    \State $\widehat{Y} \gets Q_{\theta_\ell}(h_\ell)$
    \State Compute $\mathcal{L}_{\mathrm{MAC}}$ using Eq.~\ref{eq:mac_loss} 
    \State Compute unweighted reconstruction loss $L_{\mathrm{u}}$ with unit weights for visual and text tokens
    \State $L \gets \mathcal{L}_{\mathrm{MAC}}/\operatorname{detach}(L_{\mathrm{u}})$
    \State Update $\theta_\ell$ using gradients of $L$
\EndFor
\State \Return calibrated block $Q_{\theta_\ell}$
\end{algorithmic}
\end{algorithm}

\begin{algorithm}[t]
\caption{MKR joint refinement}
\label{alg:mkr}
\begin{algorithmic}[1]
\Require FP model $F$, MAC-calibrated model $Q_{\theta}$, ROMC inputs $\mathcal{D}$
\For{each optimization iteration}
    \State Sample a minibatch $\mathcal{B}$ from $\mathcal{D}$
    \State Run $F$ and $Q_{\theta}$ on identical inputs
    \State Compute $\mathcal{L}_{\mathrm{MKR}}(x)$ for each $x\in\mathcal{B}$ using Eq.~\ref{eq:mkr_loss}
    \State $L \gets \frac{1}{|\mathcal{B}|}\sum_{x\in\mathcal{B}}\mathcal{L}_{\mathrm{MKR}}(x)$
    \State Jointly update $\theta$ across all decoder blocks using gradients of $L$
\EndFor
\State \Return refined model $Q_{\theta}$
\end{algorithmic}
\end{algorithm}

MAC calibrates blocks sequentially, aggregating reconstruction errors over the global batch. The detached denominator is constant during backpropagation. MKR computes full-vocabulary KL divergence between the final FP and quantized predictions at masked response positions, averaging first within each image and then across images. The KL temperature and loss coefficient are both one. Both stages use AdamW~\cite{adamw} with weight decay 0.01 and cosine
learning-rate decay.
Table~\ref{tab:supp-optimization} summarizes the settings:
MAC runs for 15 epochs per block and MKR for five epochs
jointly, both with a global batch size of four.

Transformation matrices and diagonal scales use learning rates
of $5\times10^{-3}$ in MAC and $1\times10^{-4}$ in MKR;
weight and activation clipping parameters use $5\times10^{-2}$
and $1\times10^{-3}$ in the corresponding stages.
We use the checkpoint from the fifth MKR epoch.

\begin{table}[t]
    \centering
    \caption{Optimization settings for MAC and MKR.}
    \label{tab:supp-optimization}
    \small
    \setlength{\tabcolsep}{4pt}
    \begin{tabular}{lcc}
        \hline
        Setting & MAC & MKR \\
        \hline
        Epochs & 15 per block & 5 \\
        Batch size & 4 & 4 \\
        Transformation / scale LR
            & $5\times10^{-3}$ & $1\times10^{-4}$ \\
        Weight / activation clipping LR
            & $5\times10^{-2}$ & $1\times10^{-3}$ \\
        \hline
    \end{tabular}
\end{table}

\subsection{Evaluation and Decoding Settings}
\label{sec:supp-evaluation}

We evaluate on seven benchmarks: MMMU~\cite{mmmu},
MME~\cite{mme}, AI2D~\cite{ai2d}, MMStar~\cite{mmstar},
POPE~\cite{pope}, SEED-Bench Image~\cite{seedbench},
and TextVQA~\cite{textvqa}.
We use each model's native prompt templates and image
preprocessing, together with the corresponding benchmark
scoring procedures. For POPE, we report pooled F1 across
its three subsets; for SEED-Bench, we evaluate image
questions only. All methods share the same evaluation
protocol within each model and attention configuration.

Evaluation uses a batch size of one per GPU and a decoding
temperature of zero. We follow the official LaViDa~\cite{lavida}
and LLaDA-V~\cite{lladav} evaluation settings, retaining each model's native
token-selection and commitment policy.
LaViDa-LLaDA and LaViDa-Dream use a response-token budget
of 16 across all benchmarks.
LLaDA-V uses 16 tokens and 16 denoising steps for TextVQA,
and two tokens and two steps for the remaining benchmarks.

For both LaViDa models, we evaluate Full-DLM and Prefix-DLM.
Full-DLM uses bidirectional attention over the entire
sequence without prefix KV caching.
Prefix-DLM restricts image--prompt prefix tokens to
attending within the prefix, allowing their keys and
values to be cached across denoising steps, while response
tokens can attend to the full sequence.
The two configurations share identical model weights,
quantization parameters, and other evaluation settings;
only the attention mask and prefix KV caching differ.
No additional calibration is performed for Prefix-DLM.
LLaDA-V retains its native bidirectional attention.


\begin{table*}[t!]
\scriptsize
\centering
\caption{
Performance comparison on LLaDA-V-8B, LaViDa-LLaDA-8B, and
LaViDa-Dream-7B across seven multimodal benchmarks.
Both Full-DLM and Prefix-DLM configurations are evaluated for LaViDa.
All quantized models use W4A4KV16.
Avg.\ denotes the arithmetic mean of per-benchmark score ratios
to the corresponding full-precision model, expressed as a percentage. DLLMQuant$^{++}$ denotes the QuaRot-based variant~\cite{dllmquant}.
Results marked with $^{\ast}$ are from our paper-based reimplementations.
Bold indicates the best reported result among quantized methods
within each configuration. Dashes indicate unreported results.
}
\label{tab:supp-baseline-comparison}
\vspace{-1mm}

\resizebox{0.96\textwidth}{!}{%
\setlength{\tabcolsep}{7pt}%
\begin{tabular}{l|ccccccc|c}
\toprule
\textbf{Method} &
\shortstack{\textbf{MMMU} \\ Acc. $\uparrow$} &
\shortstack{\textbf{MME} \\ P+C $\uparrow$} &
\shortstack{\textbf{AI2D} \\ Acc. $\uparrow$} &
\shortstack{\textbf{MMStar} \\ Avg. $\uparrow$} &
\shortstack{\textbf{POPE} \\ F1 $\uparrow$} &
\shortstack{\textbf{TextVQA} \\ VQA Score $\uparrow$} &
\shortstack{\textbf{SEED-I} \\ Acc. $\uparrow$} &
\shortstack{\textbf{Avg.} \\ Rel. (\%) $\uparrow$} \\
\midrule

\rowcolor{gray!20}
\multicolumn{9}{c}{\textit{LLaDA-V-8B}} \\
\midrule
Original (FP16)
    & 49.00 & 1994.99 & 77.78 & 60.37
    & 86.86 & 64.36 & 74.78 & 100.00 \\
\midrule
QuaRot \textcolor{gray}{(NeurIPS 2024)}
    & 47.44 & \textbf{1975.48} & 77.30 & 57.25
    & 86.60 & 60.27 & 74.17 & 97.51 \\
SpinQuant \textcolor{gray}{(ICLR 2025)}
    & 45.67 & 1870.63 & 75.84 & 57.34
    & 86.49 & 59.38 & 74.21 & 95.79 \\
FlatQuant \textcolor{gray}{(ICML 2025)}
    & 46.78 & 1965.83 & 77.10 & 58.87
    & 86.70 & \textbf{63.38} & 74.43 & 98.35 \\
MBQ \textcolor{gray}{(CVPR 2025)}
    & 22.78 & 110.28 & 8.19 & 4.05
    & 65.13 & 0.00 & 4.31 & 21.43 \\
ImpQuant \textcolor{gray}{(ICML 2026)}
    & 35.00 & 1403.92 & 57.48 & 43.87
    & 75.49 & 0.85 & 67.65 & 66.72 \\
DLLMQuant$^{++\ast}$ \textcolor{gray}{(ICML 2026)}
    & 33.11 & 803.72 & 49.71 & 37.18
    & 14.84 & 6.18 & 46.20 & 45.97 \\
FAIR-Calib$^{\ast}$ \textcolor{gray}{(ICML 2026)}
    & 47.56 & 1962.98 & 76.13 & 58.49
    & \textbf{87.17} & 61.95 & 74.57 & 98.08 \\
\rowcolor[RGB]{232,225,245}
CommitQuant (ours)
    & \textbf{48.11} & 1958.55 & \textbf{77.33} & \textbf{59.41}
    & 86.98 & 63.07 & \textbf{74.63} & \textbf{98.87} \\
\midrule

\rowcolor{gray!20}
\multicolumn{9}{c}{\textit{LaViDa-LLaDA-8B (Full-DLM)}} \\
\midrule
Original (BF16)
    & 44.33 & 1722.44 & 69.27 & 45.72
    & 83.41 & 58.84 & 66.42 & 100.00 \\
\midrule
QuaRot \textcolor{gray}{(NeurIPS 2024)}
    & 42.22 & 1682.27 & 68.56 & 46.16
    & 82.94 & 57.78 & 66.04 & 98.56 \\
SpinQuant \textcolor{gray}{(ICLR 2025)}
    & 42.11 & 1642.67 & 68.36 & 46.20
    & 82.65 & 57.92 & 65.39 & 98.01 \\
FlatQuant \textcolor{gray}{(ICML 2025)}
    & 43.11 & 1624.46 & 69.13 & 46.11
    & 83.13 & \textbf{58.38} & 66.30 & 98.70 \\
MBQ \textcolor{gray}{(CVPR 2025)}
    & 25.33 & 370.68 & 8.45 & 3.16
    & 41.01 & 5.38 & 11.04 & 24.67 \\
ImpQuant \textcolor{gray}{(ICML 2026)}
    & 34.22 & 1348.35 & 53.82 & 37.49
    & 81.55 & 43.86 & 59.21 & 82.38 \\
DLLMQuant$^{++\ast}$ \textcolor{gray}{(ICML 2026)}
    & 36.00 & 1520.42 & 64.57 & 42.93
    & \textbf{84.34} & 44.98 & 59.01 & 89.00 \\
FAIR-Calib$^{\ast}$ \textcolor{gray}{(ICML 2026)}
    & 44.00 & 1628.99 & 69.27 & 46.27
    & 83.86 & 57.85 & 66.66 & 99.18 \\
\rowcolor[RGB]{232,225,245}
CommitQuant (ours)
    & \textbf{44.22} & \textbf{1686.55} & \textbf{69.27} & \textbf{46.34}
    & 83.53 & 58.27 & \textbf{66.98} & \textbf{99.86} \\
\midrule

\rowcolor{gray!20}
\multicolumn{9}{c}{\textit{LaViDa-LLaDA-8B (Prefix-DLM)}} \\
\midrule
Original (BF16)
    & 44.22 & 1694.40 & 70.08 & 46.87
    & 83.75 & 56.33 & 66.61 & 100.00 \\
\midrule
QuaRot \textcolor{gray}{(NeurIPS 2024)}
    & 41.89 & 1691.96 & 67.94 & 45.46
    & 83.36 & 54.80 & 66.89 & 97.97 \\
SpinQuant \textcolor{gray}{(ICLR 2025)}
    & 41.00 & 1688.22 & 67.26 & 43.75
    & 83.62 & 54.50 & 65.42 & 96.64 \\
FlatQuant \textcolor{gray}{(ICML 2025)}
    & 42.78 & 1641.03 & 68.78 & \textbf{45.52}
    & \textbf{84.13} & 55.42 & 66.54 & 98.23 \\
MBQ \textcolor{gray}{(CVPR 2025)}
    & 24.89 & 378.20 & 3.79 & 1.66
    & 46.84 & 1.98 & 4.73 & 22.01 \\
ImpQuant \textcolor{gray}{(ICML 2026)}
    & 33.33 & 1429.88 & 51.49 & 39.89
    & 80.55 & 38.59 & 59.28 & 81.72 \\
DLLMQuant$^{++\ast}$ \textcolor{gray}{(ICML 2026)}
    & 37.22 & 1531.88 & 63.41 & 41.89
    & 84.02 & 42.92 & 59.35 & 88.58 \\
FAIR-Calib$^{\ast}$ \textcolor{gray}{(ICML 2026)}
    & \textbf{44.00} & 1633.65 & 68.62 & 44.39
    & 83.21 & \textbf{56.11} & 66.71 & 98.24 \\
\rowcolor[RGB]{232,225,245}
CommitQuant (ours)
    & 43.22 & \textbf{1702.46} & \textbf{69.14} & 45.41
    & 84.00 & 55.61 & \textbf{66.92} & \textbf{99.03} \\
\midrule

\rowcolor{gray!20}
\multicolumn{9}{c}{\textit{LaViDa-Dream-7B (Full-DLM)}} \\
\midrule
Original (BF16)
    & 42.33 & 1839.15 & 68.43 & 46.21
    & 85.91 & 57.45 & 69.75 & 100.00 \\
\midrule
QuaRot \textcolor{gray}{(NeurIPS 2024)}
    & 40.33 & 1805.76 & 67.68 & \textbf{45.96}
    & 85.12 & 56.38 & 68.79 & 98.24 \\
SpinQuant \textcolor{gray}{(ICLR 2025)}
    & 40.78 & 1778.91 & 68.01 & 44.52
    & 85.13 & 54.99 & 68.76 & 97.45 \\
FlatQuant \textcolor{gray}{(ICML 2025)}
    & 40.67 & 1768.59 & \textbf{68.48} & 45.91
    & \textbf{86.20} & 56.32 & 69.26 & 98.48 \\
MBQ \textcolor{gray}{(CVPR 2025)}
    & 26.11 & 733.00 & 4.95 & 6.41
    & 51.32 & 5.28 & 7.75 & 28.95 \\
ImpQuant \textcolor{gray}{(ICML 2026)}
    & 31.44 & 1504.83 & 55.31 & 38.66
    & 79.57 & 34.64 & 61.49 & 80.24 \\
DLLMQuant$^{++\ast}$ \textcolor{gray}{(ICML 2026)}
    & 33.67 & 1506.68 & 62.18 & 42.16
    & 84.57 & 35.98 & 62.85 & 84.96 \\
FAIR-Calib$^{\ast}$ \textcolor{gray}{(ICML 2026)}
    & 41.22 & 1763.29 & 67.88 & 45.37
    & 86.15 & 54.54 & \textbf{69.75} & 97.98 \\
\rowcolor[RGB]{232,225,245}
CommitQuant (ours)
    & \textbf{41.56} & \textbf{1838.74} & 67.81 & 45.19
    & 85.77 & \textbf{56.71} & 69.39 & \textbf{99.01} \\
\midrule

\rowcolor{gray!20}
\multicolumn{9}{c}{\textit{LaViDa-Dream-7B (Prefix-DLM)}} \\
\midrule
Original (BF16)
    & 42.33 & 1856.16 & 68.85 & 45.30
    & 85.84 & 57.03 & 69.73 & 100.00 \\
\midrule
QuaRot \textcolor{gray}{(NeurIPS 2024)}
    & 39.78 & 1756.27 & 67.00 & 43.55
    & 85.44 & 56.31 & 68.85 & 97.01 \\
SpinQuant \textcolor{gray}{(ICLR 2025)}
    & 40.56 & 1747.62 & 66.68 & 43.45
    & 84.96 & 54.39 & 68.62 & 96.50 \\
FlatQuant \textcolor{gray}{(ICML 2025)}
    & 42.11 & 1754.16 & \textbf{68.43} & 44.60
    & \textbf{86.20} & 56.17 & 69.37 & 98.60 \\
MBQ \textcolor{gray}{(CVPR 2025)}
    & 26.56 & 324.15 & 1.17 & 1.05
    & 57.54 & 3.13 & 2.09 & 22.82 \\
ImpQuant \textcolor{gray}{(ICML 2026)}
    & 28.78 & 1511.64 & 53.30 & 29.41
    & 78.85 & 32.32 & 57.44 & 74.67 \\
DLLMQuant$^{++\ast}$ \textcolor{gray}{(ICML 2026)}
    & 35.22 & 1465.48 & 62.08 & 40.54
    & 84.32 & 34.77 & 63.31 & 84.54 \\
FAIR-Calib$^{\ast}$ \textcolor{gray}{(ICML 2026)}
    & 41.33 & 1798.95 & 68.07 & 44.48
    & 86.00 & 56.23 & 69.37 & 98.55 \\
\rowcolor[RGB]{232,225,245}
CommitQuant (ours)
    & \textbf{42.22} & \textbf{1817.04} & 68.20 & \textbf{44.81}
    & 85.73 & \textbf{56.46} & \textbf{69.44} & \textbf{99.15} \\
\bottomrule
\end{tabular}%
}
\vspace{-1mm}
\end{table*}

\section{Additional Experimental Results}
\label{sec:supp-results}

\subsection{Comparison with MLLM and dLLM Quantization Methods}
\label{sec:supp-baselines}

Table~\ref{tab:supp-baseline-comparison} extends our comparison of QuaRot~\cite{quarot}, SpinQuant~\cite{spinquant}, and FlatQuant~\cite{flatquant} to MLLM quantization methods, MBQ~\cite{mbq} and ImpQuant~\cite{impquant}, and dLLM quantization methods, DLLMQuant~\cite{dllmquant} and FAIR-Calib~\cite{faircalib}. We evaluate these methods on LLaDA-V, LaViDa-LLaDA, and LaViDa-Dream under W4A4KV16, including both Full-DLM and Prefix-DLM configurations for LaViDa. All methods use the same pool of 128 COCO image--caption pairs for multimodal calibration and follow the same benchmark-specific evaluation protocols within each model and decoding configuration. The Avg.\ column reports the mean of the per-benchmark score ratios to the corresponding full-precision model, expressed as a percentage.

As official implementations of DLLMQuant and FAIR-Calib were unavailable at the time of our experiments, we independently reimplemented both methods based on their respective papers. Before extending them to multimodal models, we checked their implementations through text-only dLLM evaluations on PIQA~\cite{piqa} and GSM8K~\cite{gsm8k}. For the multimodal adaptations, we retained the original quantization mechanisms and the quantization hyperparameters explicitly reported in each paper, apart from the calibration-data changes described here. Implementation details and numerical settings not specified in the papers were selected by us and fixed for the reported experiments. Accordingly, the results marked with an asterisk are obtained from our paper-based implementations adapted to multimodal models, rather than results reported by the original authors.

For DLLMQuant~\cite{dllmquant}, we implement the variant based on QuaRot~\cite{quarot}, DLLMQuant$^{++}$, retaining its three components: Temporal-Mask Adaptive Sampling (TMAS), Interaction-Aware Activation Quantization (IA-AQ), and Certainty-Guided Quantization (CGQ). TMAS constructs calibration states covering different diffusion timesteps and masking ratios. IA-AQ uses attention interactions to guide activation quantization, while CGQ incorporates mask status and token confidence into Hessian-based weight-quantization error compensation. In our multimodal adaptation, the shared COCO image--caption pairs provide the image-conditioned inputs for calibration-state collection.

For FAIR-Calib~\cite{faircalib}, we preserve its two-stage procedure. In Stage~I, \emph{teacher probing}, the full-precision teacher processes 512 questions from the GSM8K training split using random commits, a 256-token generation window, and 256 diffusion steps. The resulting fixed position prior combines frontier-hit information with masked-stage teacher reliability. In Stage~II, \emph{static weight calibration}, we replace the text-only calibration corpus with the shared 128 COCO image--caption pairs and optimize the layer-wise calibration parameters through off-policy teacher forcing with a position-weighted hidden-state MSE. The Stage~I prior remains fixed during this optimization. Thus, FAIR-Calib uses the same multimodal calibration pool as the other methods, together with an additional text-only probing set for estimating its prior.

MBQ and ImpQuant exhibit substantial performance degradation under W4A4 in the evaluated multimodal diffusion models. For example, their average relative performance on LLaDA-V is 21.43\% and 66.72\%, respectively. Severe degradation of MBQ under W4A4 has also been reported for autoregressive VLMs in SplitQ~\cite{splitq}. Our adapted DLLMQuant implementation likewise exhibits pronounced degradation in the multimodal setting. On LLaDA-V, it retains only 45.97\% of full-precision performance on average, with particularly large losses on POPE and TextVQA. These results indicate that retaining the original dLLM quantization mechanisms does not, by itself, ensure effective transfer to multimodal diffusion models under our adaptation protocol.

FAIR-Calib~\cite{faircalib} preserves substantially more performance, but the advantage over FlatQuant reported for text-only dLLMs does not translate into a consistent improvement in our multimodal evaluation. On LLaDA-V, FAIR-Calib achieves 98.08\% average relative performance, compared with 98.35\% for FlatQuant. It improves MMMU and POPE over FlatQuant but scores lower on AI2D and TextVQA. On LaViDa-Dream under Full-DLM, its remaining gap to CommitQuant is particularly evident on MME and TextVQA, with deficits of 75.45 and 2.17 points, respectively.

CommitQuant achieves the highest reported average relative performance across all five configurations. Its per-benchmark lead is less widespread on LaViDa-Dream under Full-DLM: it ranks first on three of seven benchmarks, compared with five (including one tie) on LaViDa-LLaDA under Full-DLM. In the Dream Full-DLM setting, other methods lead on AI2D, MMStar, POPE, and SEED-I. Under Prefix-DLM, CommitQuant leads on three benchmarks for LaViDa-LLaDA and five for LaViDa-Dream.

\begin{figure*}[!t]
    \centering
    \includegraphics[width=\textwidth]{
        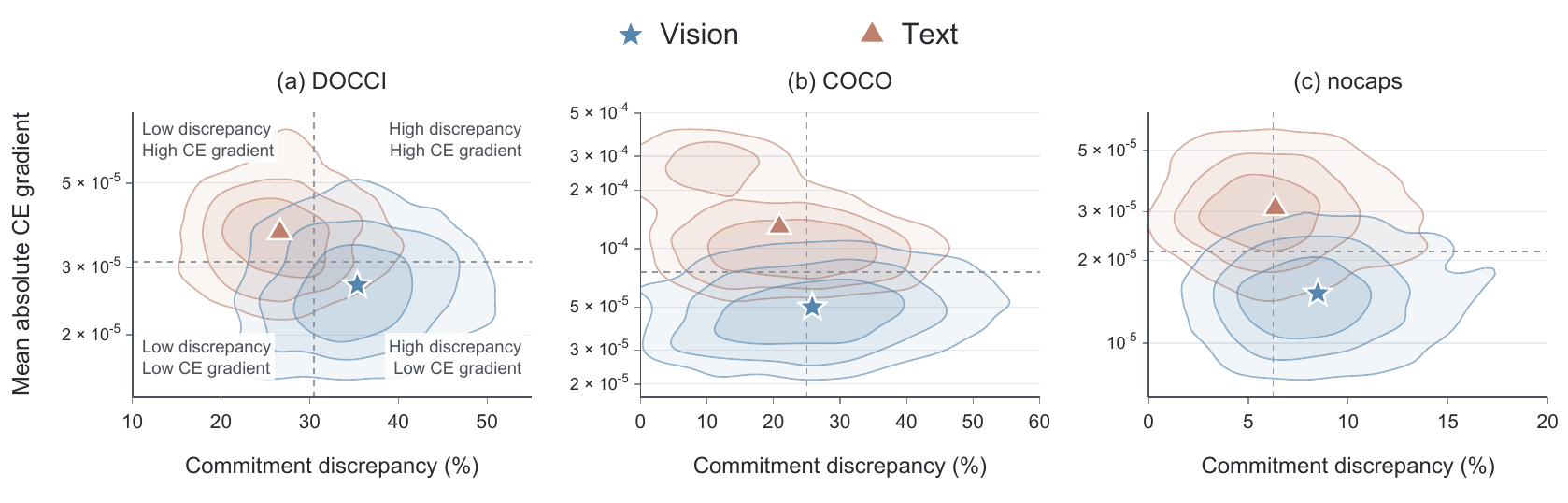
    }
    \caption{
        \textbf{CE gradients and commitment discrepancy across datasets.}
        Results on LaViDa-LLaDA-8B using five randomly sampled sets
        of 128 images from each dataset.
        Contours show per-image measurement distributions;
        stars and triangles denote modality means.
        Across all three datasets, visual features exhibit smaller
        mean absolute CE gradients but greater mean commitment
        discrepancy than textual prefix features.
    }
    \vspace{-3mm}
    \label{fig:supp_modality_sensitivity}
\end{figure*}

\subsection{Ablation Studies}
\label{sec:supp-ablation}

\paragraph{Sensitivity to calibration sample selection.}


\begin{table}[t]
    \centering
    \footnotesize
    \setlength{\tabcolsep}{3pt}
    \caption{
        Sensitivity to calibration set selection on LaViDa-LLaDA.
        Scores are averaged over MMMU, MME, MMStar, AI2D,
        and RealWorldQA after normalization to FP ($100\%$).
        $\alpha$ denotes the relative reconstruction weight.
        Green values show gains over FlatQuant in percentage points.
        The final row reports the mean and sample standard deviation.
    }
    \label{tab:calibration_sample_sensitivity}

    \begin{tabular*}{\columnwidth}{@{\extracolsep{\fill}}lccc@{}}
        \toprule
        Calibration Set & FlatQuant & CommitQuant (ours) & $\alpha$ \\
        \midrule
        \multicolumn{1}{@{}c}{1} & 98.70
          & \textbf{99.86} \textcolor{green!45!black}{(+1.16)}
          & 1.28 \\
        \multicolumn{1}{@{}c}{2} & 98.53
          & \textbf{99.55} \textcolor{green!45!black}{(+1.02)}
          & 1.28 \\
        \multicolumn{1}{@{}c}{3} & 99.79
          & \textbf{100.16} \textcolor{green!45!black}{(+0.37)}
          & 1.29 \\
        \multicolumn{1}{@{}c}{4} & 99.21
          & \textbf{99.63} \textcolor{green!45!black}{(+0.42)}
          & 1.27 \\
        \multicolumn{1}{@{}c}{5} & 98.16
          & \textbf{99.63} \textcolor{green!45!black}{(+1.47)}
          & 1.27 \\
        \midrule
        Mean $\pm$ SD
          & $98.88 \pm 0.63$
          & \textbf{99.77 $\pm$ 0.25}
            \textcolor{green!45!black}{(+0.89)}
          & $1.28 \pm 0.01$ \\
        \bottomrule
    \end{tabular*}
    \vspace{-2mm}
\end{table}
Table~\ref{tab:calibration_sample_sensitivity} evaluates sensitivity to calibration data selection on LaViDa-LLaDA using five random seeds, each defining a subset of 128 COCO training image--caption pairs.
Both methods use the same subset for each seed, and we independently estimate the MAC weight $\alpha$ for each calibration set.
The estimated weights remain stable at $1.28 \pm 0.01$.
CommitQuant outperforms FlatQuant across all five subsets, with an average gain of 0.89 percentage points. 
Across five benchmarks, including RealWorldQA~\cite{realworldqa},
mean FP-normalized scores are $99.77 \pm 0.25\%$ for CommitQuant
and $98.88 \pm 0.63\%$ for FlatQuant.

\vspace{-2mm}
\paragraph{Component ablation on LaViDa-Dream.}
\begin{table}[t]
    \centering
    \caption{
        Ablation of MAC and MKR on LaViDa-Dream-7B.
        All configurations use response-only masked calibration (ROMC).
        The first row reports the ROMC-only baseline.
    }
    \label{tab:dream_calibration_loss_ablation}
    \footnotesize
    \setlength{\tabcolsep}{3.5pt}
    \resizebox{\columnwidth}{!}{%
    \begin{tabular}{@{}cc|c|ccc@{}}
        \toprule
        \multicolumn{2}{c|}{\textbf{Calibration components}} &
        \textbf{MMMU} & \multicolumn{3}{c}{\textbf{MME}} \\
        \makebox[5em][c]{MAC} & \makebox[5em][c]{MKR} &
        Acc. (\%) $\uparrow$ &
        Perception $\uparrow$ &
        Cognition $\uparrow$ &
        Total $\uparrow$ \\
        \midrule
                     &              &
        41.33 & 1407.91 & 369.64 & 1777.56 \\
        $\checkmark$ &              &
        41.44 & 1436.19 & 361.79 & 1797.97 \\
                     & $\checkmark$ &
        41.48 & 1418.04 & 383.21 & 1801.25 \\
        $\checkmark$ & $\checkmark$ &
        \textbf{41.56} & \textbf{1450.53} &
        \textbf{388.21} & \textbf{1838.74} \\
        \bottomrule
    \end{tabular}%
    }
    \vspace{-2mm}
\end{table}
Table~\ref{tab:dream_calibration_loss_ablation} extends the calibration-component ablation on LaViDa-LLaDA to LaViDa-Dream-7B, using identical ROMC inputs across configurations.
MAC and MKR each improve MMMU accuracy and total MME over the ROMC baseline. Combining them achieves the best results across all reported metrics, reaching 41.56\% on MMMU and 1838.74 on MME.
In particular, adding MAC to MKR improves MME by 37.49 points. These results support the complementary benefits of the two calibration stages on the Dream-based model.

\vspace{-2mm}
\paragraph{Modality sensitivity across datasets.}
We extend the modality sensitivity analysis to COCO~\cite{coco},
DOCCI~\cite{docci}, and nocaps~\cite{nocaps} using LaViDa-LLaDA-8B.
For each dataset, we randomly sample five subsets of
128 images each.
For each image, we measure mean absolute CE gradients of visual
and textual prefix features and commitment discrepancy under
modality-specific quantization-error injection, following
Section~\ref{sec:supp-discrepancy}.
Both interventions use the same FP decoding states.
Figure~\ref{fig:supp_modality_sensitivity} shows the per-image
distributions.

Across all three datasets, visual features exhibit smaller
mean absolute CE gradients than textual features,
yet visual quantization-error injections produce greater
mean commitment discrepancy.
Thus, the modality ordering suggested by CE-gradient
magnitude is reversed when sensitivity is measured through
actual changes in commitment decisions.
The consistency of this pattern across datasets indicates
that the mismatch observed on COCO is not confined to that
dataset.
These results further support measuring commitment
discrepancy directly when determining the relative
reconstruction priority of visual and textual features.

\subsection{Additional Qualitative Comparisons}
\label{sec:supp-qualitative}
\begin{figure*}[p]
    \centering
    \includegraphics[width=0.99\textwidth]{
        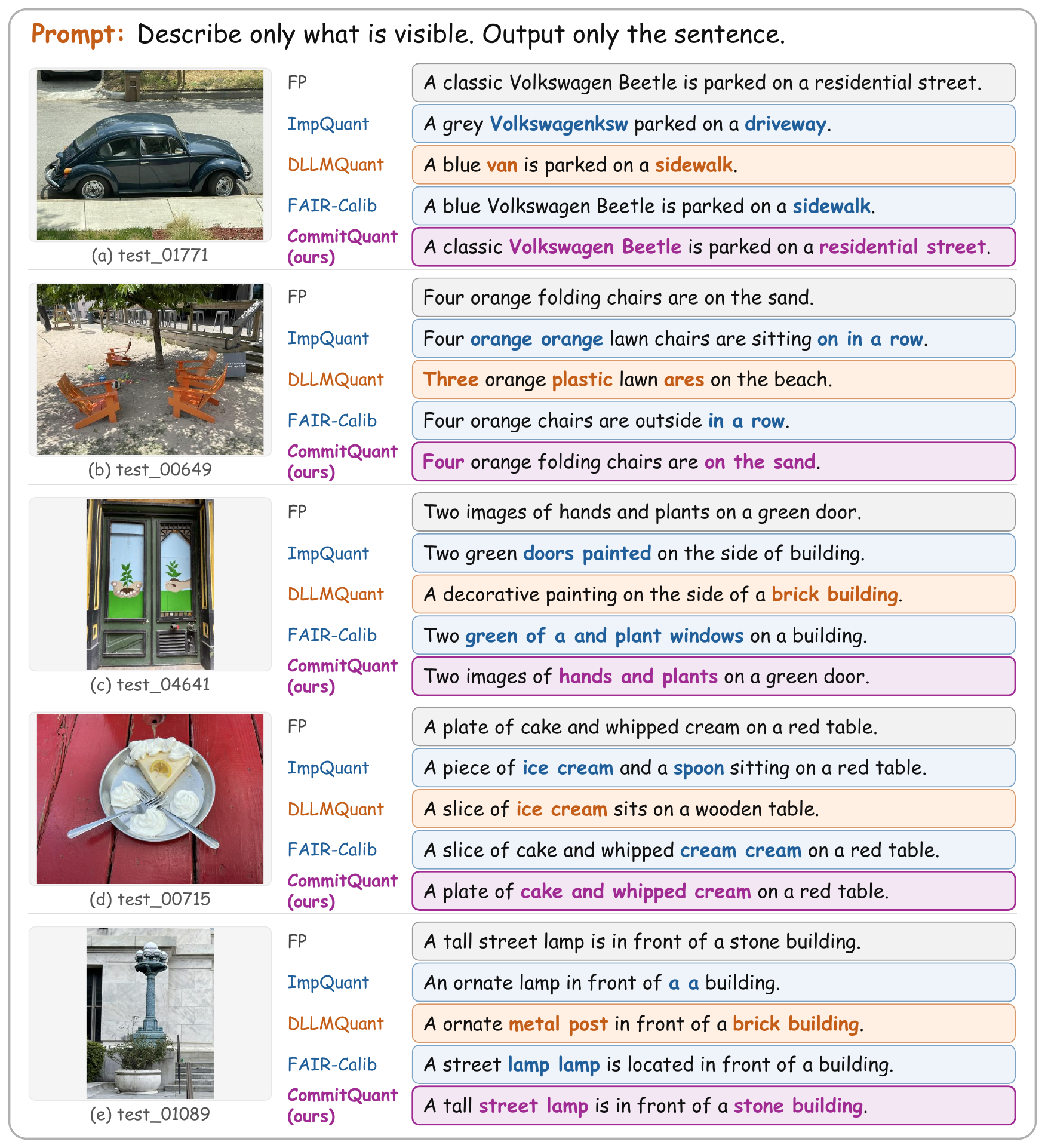
    }
    \caption{
        \textbf{Qualitative comparison on DOCCI.}
        Responses from FP, ImpQuant, DLLMQuant, FAIR-Calib,
        and CommitQuant on five selected DOCCI images,
        evaluated using LaViDa-LLaDA-8B under Full-DLM
        with a 32-token generation budget and 16 denoising steps.
        All responses are shown verbatim.
        Colored phrases highlight selected descriptive
        differences or word repetitions.
        CommitQuant reproduces the FP response in all five examples,
        illustrating preservation of FP descriptions.
    }
    \label{fig:docci_qualitative}
\end{figure*}

We compare FP, ImpQuant, DLLMQuant, FAIR-Calib, and CommitQuant on selected DOCCI~\cite{docci} images using LaViDa-LLaDA-8B. All methods share the same prompt and decoding settings within each example. Figure~\ref{fig:docci_qualitative} presents five short-response examples generated with a 32-token budget and 16 denoising steps. We instruct the model to ``Describe only what is visible. Output only the sentence.'' CommitQuant reproduces the FP response in all five examples. For example, DLLMQuant changes the Beetle to a ``van'' and the chair count from four to three. ImpQuant and DLLMQuant describe the cake as ``ice cream,'' while FAIR-Calib retains the main food description but repeats ``cream.''

\begin{figure*}[p]
    \centering
    \includegraphics[
        width=\textwidth,
        height=0.99\textheight,
        keepaspectratio
    ]{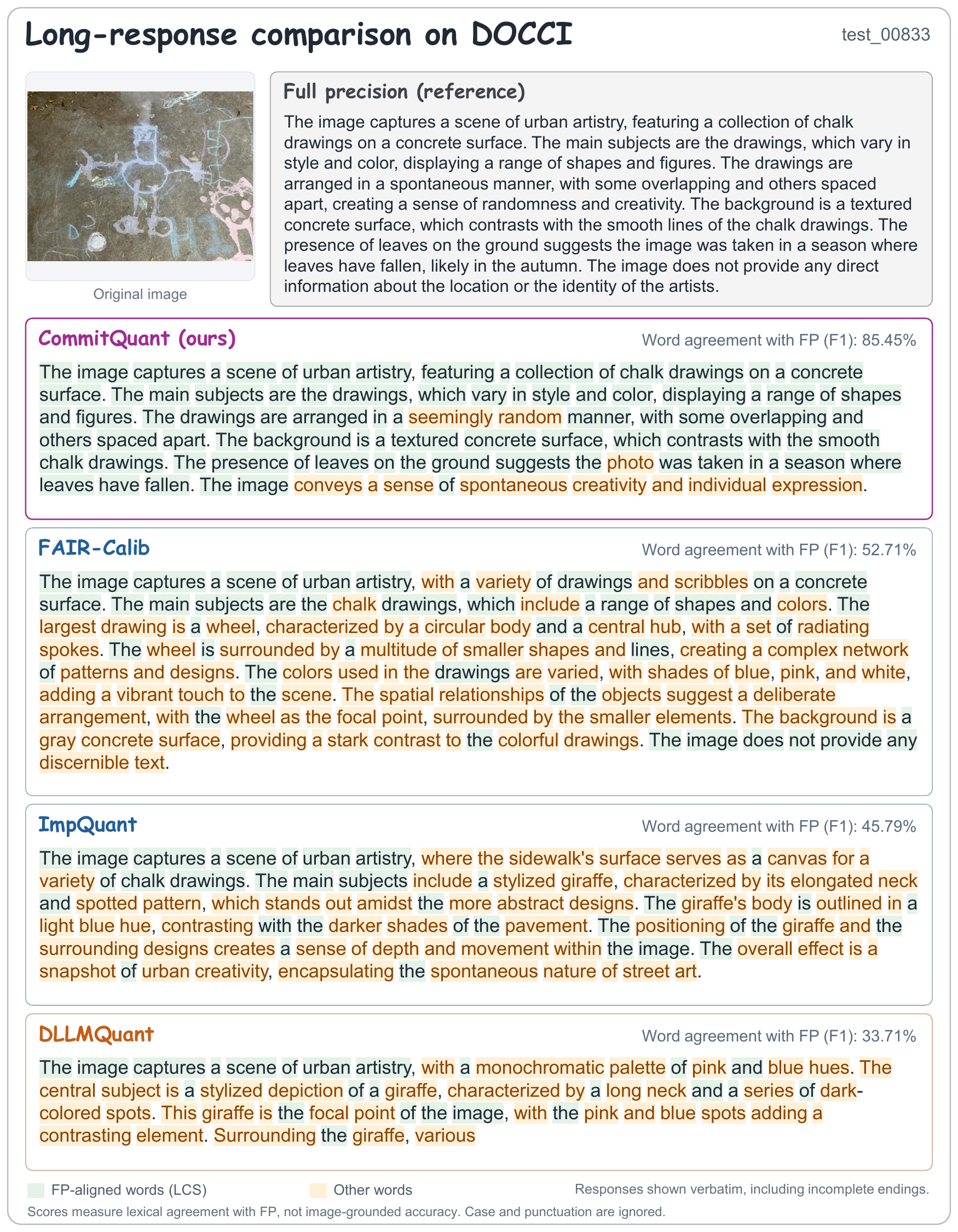}
    \vspace{-3mm}
    \caption{
        \textbf{Long-response agreement with FP.}
        Word-level agreement with the FP reference on DOCCI
        sample \texttt{test\_00833}, evaluated using
        LaViDa-LLaDA-8B.
    }
    \label{fig:docci_long_00833}
\end{figure*}

\begin{figure*}[p]
    \centering
    \includegraphics[
        width=\textwidth,
        height=0.99\textheight,
        keepaspectratio
    ]{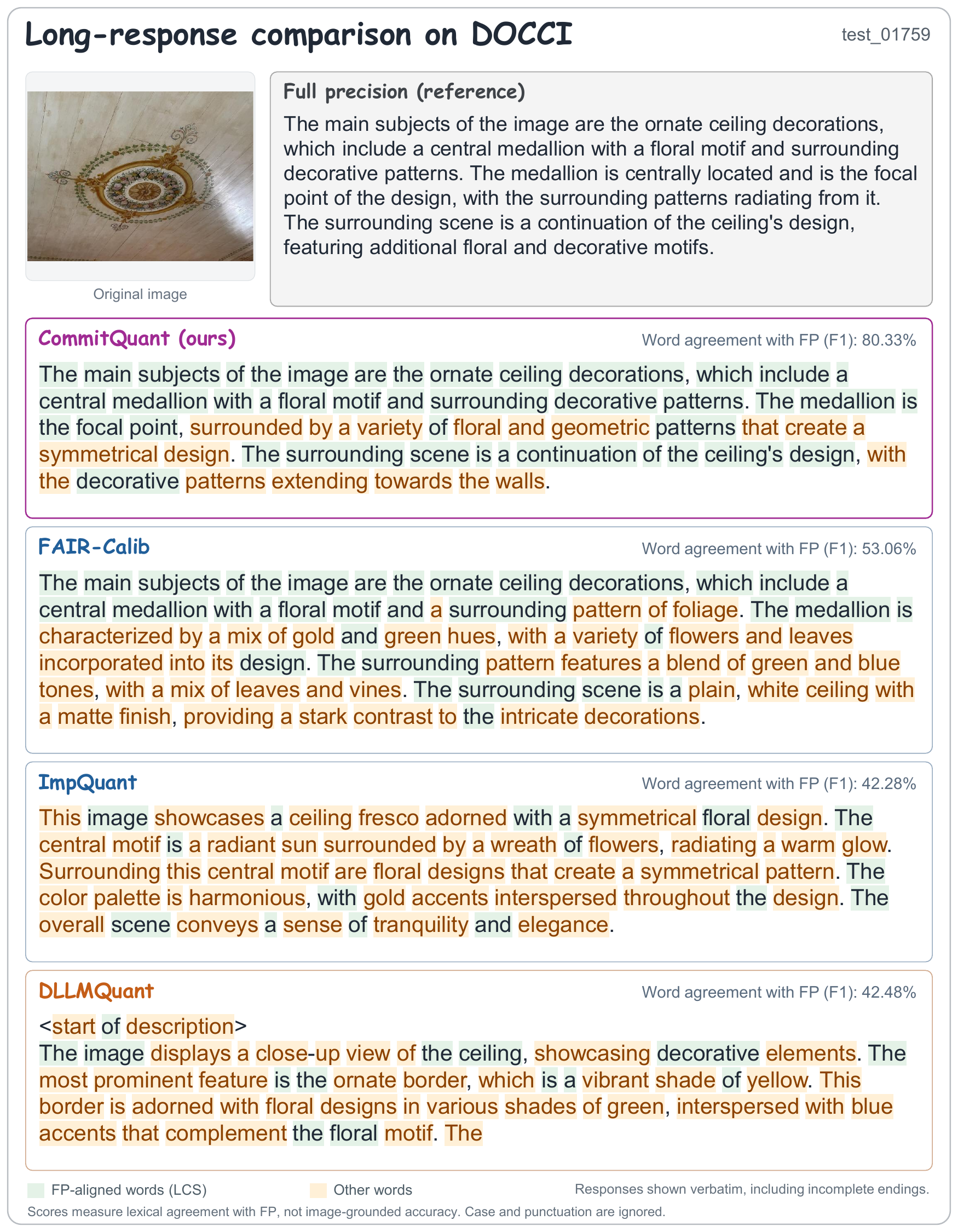}
    \vspace{-3mm}
    \caption{
        \textbf{Long-response agreement with FP.}
        Word-level agreement with the FP reference on DOCCI
        sample \texttt{test\_01759}, evaluated using
        LaViDa-LLaDA-8B.
    }
    \label{fig:docci_long_01759}
\end{figure*}

We further examine longer responses in Figs.~\ref{fig:docci_long_00833} and~\ref{fig:docci_long_01759}. For the ceiling example, we use 128 tokens and 128 denoising steps with the prompt: \emph{``Describe the visible image in three complete sentences totaling 50--80 words. In the first sentence identify the main subjects. In the second describe their visible appearance and spatial arrangement. In the third describe the surrounding scene. Do not invent details or identify a person, brand, or location unless clearly visible. Output only the caption.''} For the chalk-drawing example, we use 256 tokens and 256 denoising steps with the prompt: \emph{``Describe the image in one detailed paragraph
of approximately 100--140 words, using complete
sentences. Start by identifying the main subject
or objects. Describe their visible attributes,
including colors, shapes, counts, and actions,
then explain their spatial relationships and the
surrounding scene. Include distinctive background
details and clearly readable text when relevant.
Describe only what is supported by the image;
do not guess identities, locations, or obscured text.
Output only the description, with no heading or
introduction, and finish with a complete sentence.''}
CommitQuant achieves the highest word-level F1 agreement with FP in both examples (80.33\% and 85.45\%). Responses are shown verbatim; lexical agreement with FP does not establish image-grounded factual accuracy.

\end{document}